\documentclass[letterpaper, 10 pt, journal, twoside]{ieeetran}

\IEEEoverridecommandlockouts                              

\usepackage{amsmath} 
\usepackage{amssymb}  

\usepackage[pdftex]{graphicx}
\usepackage{url}
\usepackage{makecell}

\newcommand{\figref}[1]{{Fig.~\ref{#1}}}
\newcommand{\bm}[1]{\mbox{\boldmath{$#1$}}}

\title{TACT: Humanoid Whole-body Contact Manipulation through Deep Imitation Learning with Tactile Modality}

\author{Masaki Murooka$^{1}$, Takahiro Hoshi$^{2}$, Kensuke Fukumitsu$^{2}$, Shimpei Masuda$^{1,3}$, \\Marwan Hamze$^{2}$, Tomoya Sasaki$^{2}$, Mitsuharu Morisawa$^{1}$, and Eiichi Yoshida$^{2}$
\thanks{Manuscript received: March, 2, 2025; Revised May, 14, 2025; Accepted June, 5, 2025.}
\thanks{This paper was recommended for publication by Editor Olivier Stasse upon evaluation of the Associate Editor and Reviewers' comments.
  This work was supported in part by JSPS KAKENHI Grant Number 22H05002 and 22K17984.} 
\thanks{$^{1}$Masaki Murooka, Shimpei Masuda, and Mitsuharu Morisawa are with
  CNRS-AIST JRL (Joint Robotics Laboratory), IRL and
  National Institute of Advanced Industrial Science and Technology (AIST),
  1-1-1 Umezono, Tsukuba, Ibaraki 305-8560, Japan.
  {\tt\small \{m-murooka, masuda.shimpei, m.morisawa\}@aist.go.jp}}%
\thanks{$^{2}$Takahiro Hoshi, Kensuke Fukumitsu, Marwan Hamze, Tomoya Sasaki, and Eiichi Yoshida are with
  Tokyo University of Science,
  6-3-1 Niijuku, Katsushika-ku, Tokyo 125-8585, Japan.
  {\tt\small \{marwan.hamze, tomoya, eiichi.yoshida\}@rs.tus.ac.jp}}%
\thanks{$^{3}$Shimpei Masuda is also with
  University of Tsukuba,
  1-1-1 Tennodai, Tsukuba, Ibaraki 305-8577 Japan.}%
\thanks{Digital Object Identifier (DOI): see top of this page.}
}

\begin{document}

\maketitle

\markboth{IEEE Robotics and Automation Letters. Preprint Version. Accepted June, 2025}
{Murooka \MakeLowercase{\textit{et al.}}: TACT: Humanoid Whole-body Manipulation through Tactile-based Imitation Learning}

\setlength{\floatsep}{8pt}
\setlength{\textfloatsep}{10pt}
\setlength{\abovecaptionskip}{4pt}
\setlength{\abovedisplayskip}{4pt}
\setlength{\belowdisplayskip}{4pt}

\begin{abstract}
  Manipulation with whole-body contact by humanoid robots offers distinct advantages, including enhanced stability and reduced load. On the other hand, we need to address challenges such as the increased computational cost of motion generation and the difficulty of measuring broad-area contact. We therefore have developed a humanoid control system that allows a humanoid robot equipped with tactile sensors on its upper body to learn a policy for whole-body manipulation through imitation learning based on human teleoperation data. This policy, named tactile-modality extended ACT (TACT), has a feature to take multiple sensor modalities as input, including joint position, vision, and tactile measurements. Furthermore, by integrating this policy with retargeting and locomotion control based on a biped model, we demonstrate that the life-size humanoid robot RHP7 Kaleido is capable of achieving whole-body contact manipulation while maintaining balance and walking. Through detailed experimental verification, we show that inputting both vision and tactile modalities into the policy contributes to improving the robustness of manipulation involving broad and delicate contact.
\end{abstract}


\section{Introduction}

\IEEEPARstart{I}{n} order for humanoid robots to perform a wide range of tasks in daily life and industrial environments, it is essential that they possess sensory-motor systems capable of responding to a variety of human-like behaviors. The majority of the locomotion and manipulation achieved by humanoid robots to date is limited to behaviors involving contacts only at the extremities such as hands and feet. In contrast, humans utilize their entire body to interact with the environment and objects, thereby evenly distributing the load and enhancing stability. Achieving such behavior necessitates advanced sensory-motor mapping capabilities to perceive whole-body contact and select appropriate actions accordingly. However, conventional model-based humanoid control methods encounter challenges with modeling and computational costs, leading to the conversion of distributed tactile measurements to low-dimensional data such as contact areas. Consequently, fine feedback by exploiting distributed tactile measurements still remains a significant challenge.

\begin{figure}[tpb]
  \centering
  \includegraphics[width=0.95\columnwidth]{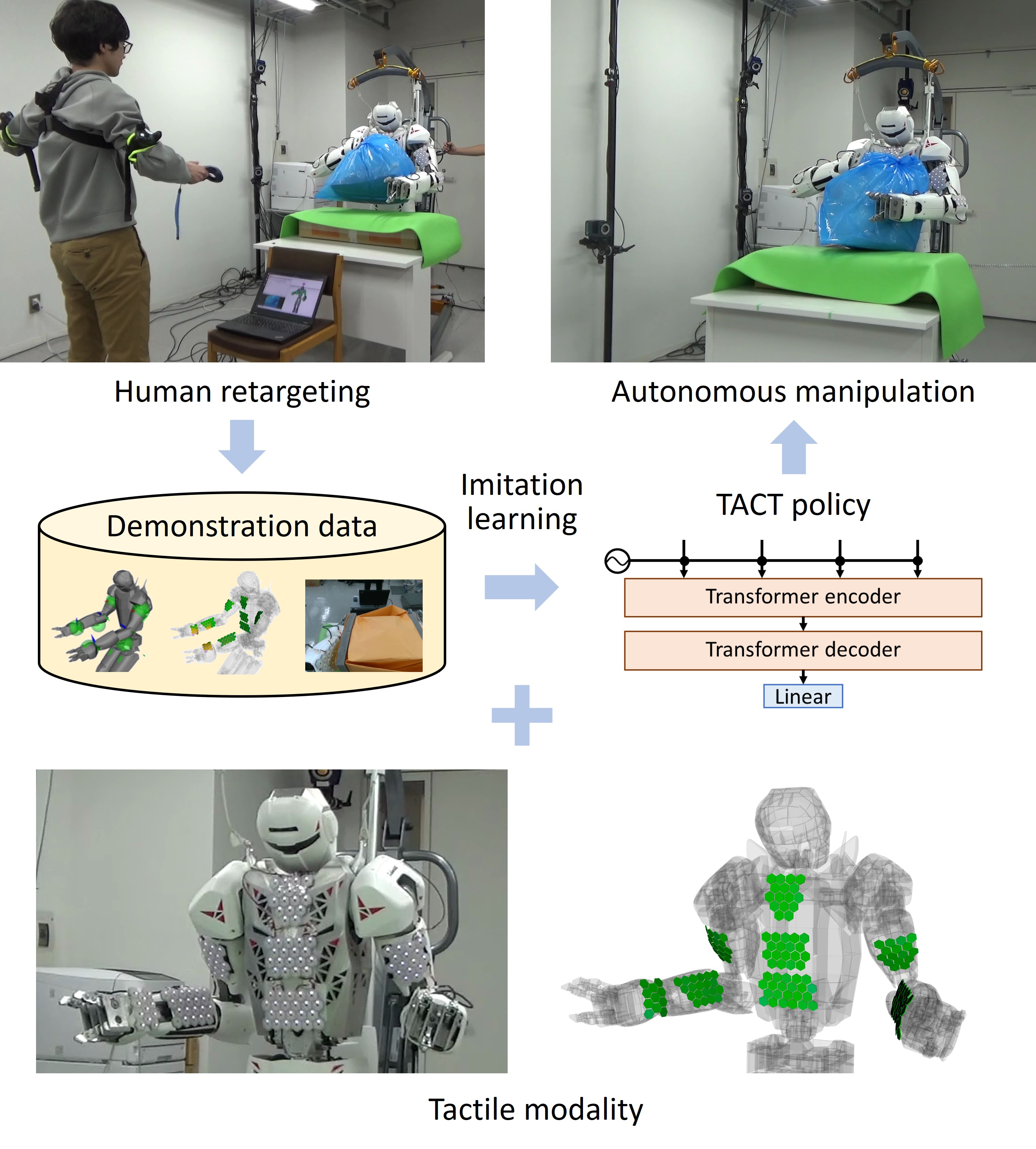}
  \caption{Humanoid manipulation through imitation learning with tactile modality.}
  \label{fig:intro}
\end{figure}

In this study, we propose a learning-based control system that enables a humanoid to achieve loco-manipulation with delicate and rich contact. This system is achieved by mounting distributed tactile sensors on the upper body and applying control based on the measurements from these sensors. The core of the control system is a Transformer-based policy that processes multimodal inputs, including tactile, visual, and proprioception (joint position) data, and generates actions for the future horizon. We extend action chunking with transformers (ACT)~\cite{ALOHA:Zhao:RSS2023}, a prominent imitation learning model for robot manipulation, and propose tactile-modality extended ACT (TACT), which can accept tactile measurements as input. The sensory-motor data necessary for model training is obtained through online teleoperation of the robot, wherein the posture of a human wearing pose trackers on the upper body is retargeted to the humanoid. Experiments with the life-size humanoid robot RHP7 Kaleido have confirmed the effectiveness of the proposed method in utilizing tactile measurements to stably hold multiple types of objects with the whole body. The proposed system is characterized by the integration of two distinct yet interconnected components: reliable model-based retargeting and locomotion control, and learning-based tactile feedback manipulation with high flexibility. This integration enables the humanoid to interact closely with objects while maintaining a stable balance.



\section{Related Works and Contributions}

This study encompasses three research domains of humanoid robotics: whole-body loco-manipulation, tactile sensing, and learning-based control. The features of this study in each domain are described below.


\subsection{Humanoid Whole-Body Loco-Manipulation}

In the majority of prior studies on humanoid loco-manipulation, robots have interacted with objects exclusively through their hands~\cite{LocomanipPlan:Jorgense:ICRA2020,LocomanipControl:Murooka:RAL2021,DynamicLocomanipMPC:Li:ACC2023,MultiContactPositionControlled:Rouxel:RAL2024}. The primary reasons for this approach are twofold: (Constraint~1) the robot directly measures and controls the forces from the object using the 6-axis force/torque sensors mounted on the wrist, and (Constraint~2) predefining the contact points maintains the simplicity of the robot's dynamics model. Nevertheless, there exist several approaches that facilitate the generation of motion for manipulation without constraining the contact points to the hands, through advancements in modeling and search, as outlined below. Some methods have been proposed for generating whole-body contact postures by solving optimization problems based on a robot kinematics model that assumes contact at any points~\cite{MulticontactManifolds:Brossette:TRO2018,WholebodyContactPosture:Murooka:RAL2020}. Other methods have been proposed for searching for robot postures involving contact at any points by random sampling in the configuration space of the robot and object~\cite{PowerGraspPlan:Roa:ICRA2012,WholebodyPush:Murooka:ICRA2015}. However, these methods are computationally intensive, making it difficult to generate dense sequences of poses in the time direction. Consequently, they are typically limited to generating poses at keyframes, and stable interpolation between keyframes remains challenging. For the stabilization control of motions involving contact with body parts other than the extremities, there are methods based on torque control~\cite{MultiContactKnee:Henze:IROS2017} and methods based on quasi-static motion that assume the robot's exact mass properties are known~\cite{MultiContact:Hiraoka:AR2021}. However, these control methods necessitate that the contact points on the robot's body be known in advance, precluding their application to our focus area of whole-body loco-manipulation, wherein contact points are determined during motion.


\subsection{Tactile Sensing-based Humanoid Control}

In order to overcome (Constraint~1) mentioned earlier, it is effective to directly measure contacts by mounting sensors on body parts of the robot where contact is expected. Several sheet-shaped distributed tactile sensors have been developed and mounted on multiple body parts of humanoids~\cite{Maiolino:CapacitiveTactileSystem:IEEESensors2013,Cheng:RobotSkin:IEEE2019}. These sensors have been applied to a variety of tasks, including whole-body manipulation of large objects~\cite{Mittendorfer:WholebodyTactileManipulation:AdvancedRobotics2015,Punyo1:Goncalves:SoftRobot2022}, whole-body multi-contact motion~\cite{TactileMultiContact:Murooka:RAL2024}, whole-body interaction with humans~\cite{DeanLeon:TactileComplianceHumanoid:ICRA2019}, in-hand manipulation~\cite{Weiner:TactileGrasp:IROS2021}, and learning-based bimanual manipulation~\cite{Shikada:TactileBimanual:IROS2024}.
Some model-based approaches incorporating tactile sensing require a geometric calibration process to construct a spatial map of the sensor cells~\cite{Chefchaouni:TactileSkinCalibration:Sensors2023}. In contrast, some learning-based methods have demonstrated effective policy learning by exploiting the spatial structure of tactile sensors using graph neural networks, for tasks such as grasping~\cite{Funabashi:GCNGrasp:RAL2022} and in-hand manipulation~\cite{Yang:GNNInHandManip:RAL2023}. In this study, we use a Transformer-based network to implicitly learn correlations among sensor cells without requiring prior calibration.


\subsection{Learning-based Humanoid Control}

In recent years, research on generating humanoid motion using learning-based models such as neural networks has grown substantially. Compared to traditional approaches that rely on hand-crafted motion models, learning-based methods offer greater flexibility in handling high-dimensional inputs and outputs, making them particularly suitable for processing distributed tactile sensor measurements. Training on data that includes contacts across various body parts is expected to relax (Constraint~2) discussed in the previous section. Learning-based humanoid control methods can be broadly categorized into reinforcement learning~\cite{ReinforcementLearningSurvey:Ibarz:IJRR2021} and imitation learning~\cite{WhatMattersImitationLearning:Mandlekar:CoRL2021}. Reinforcement learning approaches typically train policies through domain-randomized parallel simulation and transfer them to real-world robots for robust locomotion under disturbances~\cite{DigitLocomotionRL:Radosavovic:ScienceRobotics2024,HumanoidParkour:Zhuang:CoRL2024}. In contrast, imitation learning approaches often assume a preexisting walking and retargeting controller and train a policy from teleoperation data to generate commands based on sensor inputs~\cite{LocomanipImitation:Seo:Humanoids2023,ImitationLearningHRI:CardenasPerez:RAL2024}. Two-layer architectures, where an imitation-learned policy commands a reinforcement-learned locomotion controller, have also been proposed~\cite{OmniH2O:He:CoRL2024,HumanPlus:Fu:CoRL2024}. Recent studies have demonstrated reinforcement learning for loco-manipulation~\cite{LocomanipRL:Dao:ICRA2024} and imitation learning for multi-contact motion using Flow Matching~\cite{Rouxel:FlowMatchingMultiContact:Humanoids2024}. Building on this trend, our work tackles the challenging problem of whole-body contact manipulation. Due to the large search space involved, reinforcement learning is expected to be inefficient in this context. We therefore adopt an imitation learning approach that integrates tactile sensing into a Transformer-based policy.


\begin{figure*}[h]
  \centering
  \includegraphics[width=1.85\columnwidth]{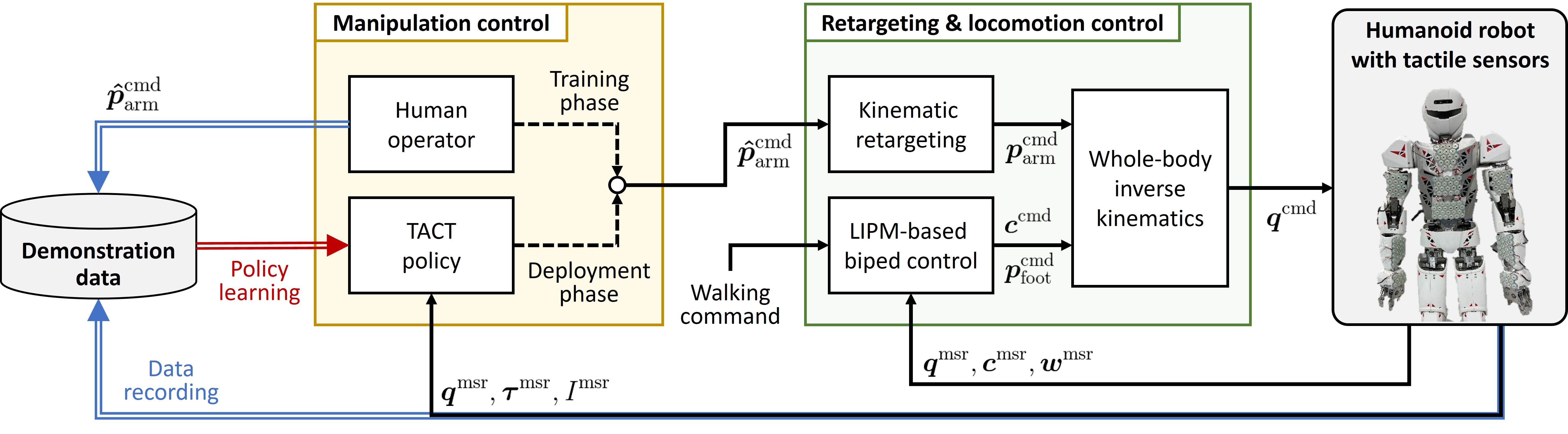}
  \caption{Humanoid control system for whole-body contact manipulation with tactile feedback.
    \newline
    \footnotesize{The control system consists of two layers: an upper layer of learning-based manipulation control and a lower layer of model-based retargeting and locomotion control. $\bm{c}$ is the CoM position, $\bm{p}$ is the pose of the feet or arms, $\bm{q}$ is the joint position, $\bm{w}$ is the contact force, $\bm{\tau}$ is the tactile measurement, and $\bm{I}$ is the camera image. The superscripts in the symbols, $\mathrm{cmd}$ and $\mathrm{msr}$, respectively represent command and measured values.}}
  \label{fig:system}
\end{figure*}

\subsection{Contributions of this Paper}

The contributions of our work are threefold: (i) we extended the major imitation learning method for robot manipulation to handle the tactile modality, which is essential for contact-rich whole-body manipulation; (ii) we integrated model-based retargeting and locomotion control with learning-based manipulation motion generation to construct a humanoid control system that provides both reliability and flexibility; and (iii) we demonstrated experiments in which a life-size humanoid robot with tactile sensors placed on its upper body achieved loco-manipulation tasks involving delicate and rich contact. To the best of our knowledge, this is the first study to achieve whole-body contact loco-manipulation using a learning-based manipulation control with a Transformer-based policy, using tactile sensors placed all over the upper body of a life-size humanoid robot.


\section{Control System Overview} \label{sec:system}

As illustrated in~\figref{fig:system}, the architecture of our control system comprises two distinct layers: an upper layer responsible for generating manipulation motions with tactile feedback, and a lower layer that handles retargeting and locomotion control. In this study, we implement a conventional model-based approach for retargeting and locomotion control, leveraging the control system that we have developed thus far~\cite{IREXFrendsDemo:Benallegue:RAM2025}. During the phase of collecting training data, a human operator wears pose trackers on the upper body and demonstrates the motions on-site. During the phase of deploying the learned policy, the policy infers motion commands in the same format as the training phase from tactile, visual, and proprioception sensor measurements. In model-based retargeting and locomotion control, it guarantees the physical constraints of the robot, while in learning-based manipulation control, it generates whole-body contact motions that require complex skills. In the following sections, retargeting and locomotion control is described in Section~\ref{sec:retarget-locomotion}, and manipulation control is described in Section~\ref{sec:manip-learn}. After that, in Section~\ref{sec:exp}, we show real-world experiments of whole-body contact manipulation by a humanoid.



\section{Retargeting and Locomotion Control} \label{sec:retarget-locomotion}

\begin{figure}[tpb]
  \centering
  \includegraphics[width=0.98\columnwidth]{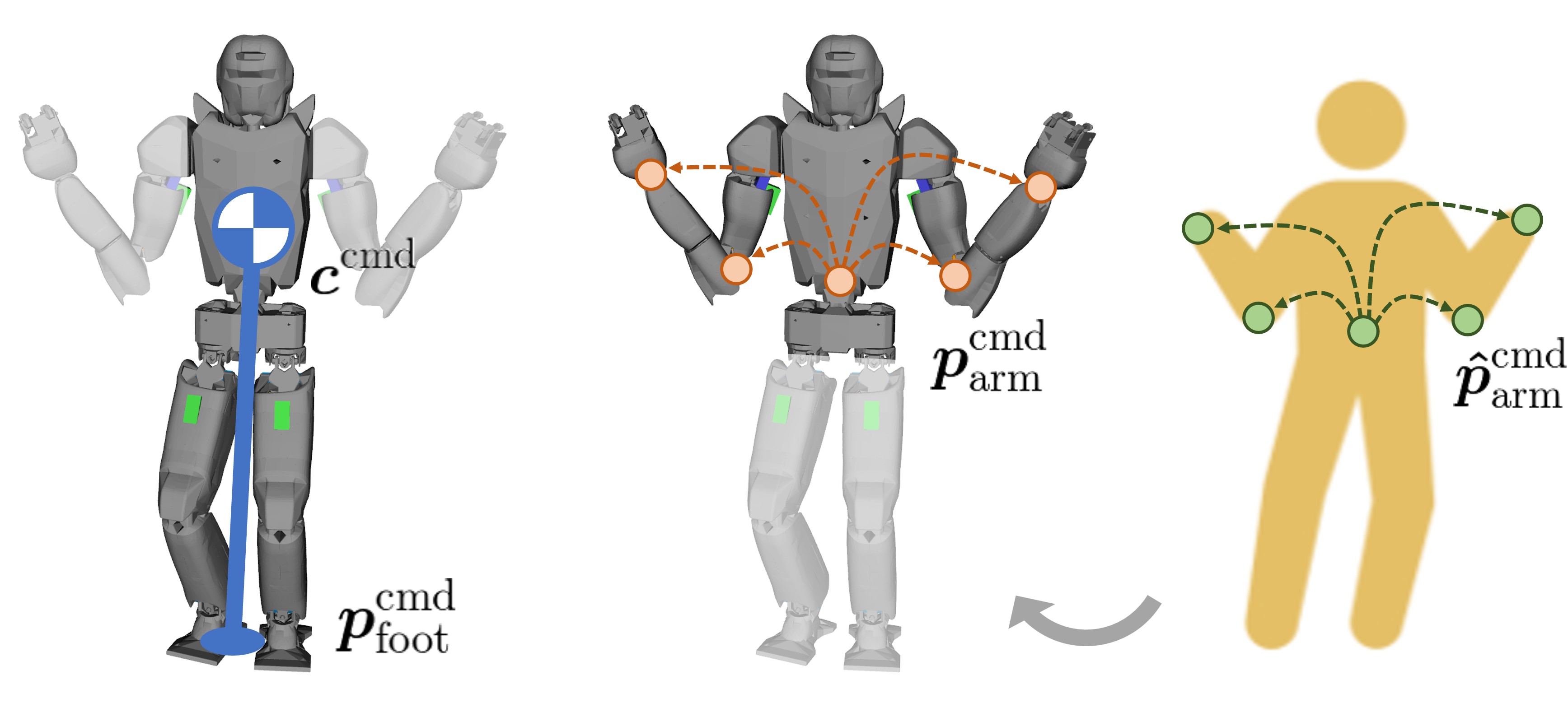}\\
  \begin{minipage}{0.42\columnwidth}
    \begin{center} \footnotesize (A) LIPM-based biped control \end{center}
  \end{minipage}
  \begin{minipage}{0.56\columnwidth}
    \begin{center} \footnotesize (B) Kinematic retargeting \end{center}
  \end{minipage}
  \caption{Retargeting and locomotion control.}
  \label{fig:retarget-locomotion}
\end{figure}

At the core of the controller lies inverse kinematics (IK) calculation, which is formulated as a quadratic programming (QP) problem. As illustrated in~\figref{fig:retarget-locomotion}, the target poses of the center of mass (CoM), feet, and arms are calculated in real time using the following components and are provided to IK.


\subsection{LIPM-based Biped Control}

We use a previously developed controller for humanoid loco-manipulation~\cite{LocomanipControl:Murooka:RAL2021}, which models the biped dynamics using the linear inverted pendulum mode (LIPM)~\cite{LIPM:Kajita:IROS2001}. This controller maintains balance even during dynamic whole-body motions induced by teleoperation or a learned policy.

It first computes the command CoM $\bm{c}^{\mathrm{cmd}}$, which is dynamically consistent with the given walking command (future footstep sequence). Discrete preview control is used to solve the following optimization problem, formulated to track the reference zero moment point (ZMP) $\bm{z}^{\mathrm{ref}}$ derived from the footstep sequence, under LIPM dynamics~\cite{PreviewControl:Kajita:ICRA2003}:
\begin{align}
  \min_{\bm{\dddot{c}}^{\mathrm{cmd}}} \ \sum_{t = 0}^{\infty} \left( \left\| \bm{z}^{\mathrm{cmd}}[t] - \bm{z}^{\mathrm{ref}}[t] \right\|^2 + k_c \left\| \bm{\dddot{c}}^{\mathrm{cmd}}[t] \right\|^2 \right)
\end{align}
The reference trajectory $\bm{z}^{\mathrm{ref}}$ is computed to pass through the support regions defined by the footstep sequence. As long as $\bm{z}^{\mathrm{cmd}}$ follows this trajectory and remains within the support region, the robot can maintain balance and avoid falling.

To further stabilize upright posture and walking under external disturbances, feedback is applied based on the error in the divergent component (DCM) $\bm{\xi}$ of the LIPM~\cite{DcmWalk:Englsberger:TRO2015}:
\begin{align}
  \bm{z}^{\mathrm{cmd\mathchar`-fb}} = \bm{z}^{\mathrm{cmd}} + k_{\mathrm{\xi}} (\bm{\xi}^{\mathrm{msr}} - \bm{\xi}^{\mathrm{cmd}})
\end{align}
The LIPM can be decomposed into stable and unstable modes; the DCM captures the latter. Providing feedback only on the DCM allows effective stabilization of biped walking with minimal control complexity.

Finally, the contact wrenches at the soles $\bm{w}_{\mathrm{foot}}^{\mathrm{cmd}}$ are regulated to match the feedback-adjusted ZMP $\bm{z}^{\mathrm{cmd\mathchar`-fb}}$. Assuming a position-controlled robot, the sole pose $\bm{p}_{\mathrm{foot}}^{\mathrm{cmd}}$ is computed using damping control, which assigns velocity to the contact point proportionally to the wrench error~\cite{LocomanipControl:Murooka:RAL2021}:
\begin{align}
  \bm{\dot{p}}_{\mathrm{foot}}^{\mathrm{cmd}} = k_{\mathrm{w}} (\bm{w}_{\mathrm{foot}}^{\mathrm{msr}} - \bm{w}_{\mathrm{foot}}^{\mathrm{cmd}})
\end{align}

\subsection{Kinematic Retargeting} \label{sec:retargeting}

In order to intuitively control the upper body of the humanoid and perform whole-body contact manipulation, the posture of the human upper body is measured and retargeted to the humanoid in real time. The human operator wears trackers that can measure 3D poses on the waist, elbows, and wrists. The target poses $\bm{p}_{\mathrm{arm}}^{\mathrm{cmd}}$ for the robot's elbows and wrists relative to the robot's waist are calculated from the measured poses $\bm{\hat{p}}_{\mathrm{arm}}^{\mathrm{cmd}}$ for the human elbows and wrists relative to the human waist.

To account for variations in operator body proportions, a calibration procedure is performed before teleoperation. The operator holds the arm in three reference postures: forward, sideways, and upward. For each posture, the elbow-wrist tracker pair defines one of three orthogonal axes, and their intersection is treated as the shoulder origin. The lengths of the upper and lower arm segments are then measured and compared with those of the robot to compute scaling factors used for accurate pose mapping.

The responsiveness of the tracking is controlled by updating the IK target poses using a first-order lag system. In our setup, the time constant is set to approximately 0.16~s, which achieves smooth motion while avoiding excessive sensitivity to small fluctuations in the operator's movements. In addition, model-based admittance control was implemented for selected body parts to enhance compliance during contact, which is particularly important for position-controlled humanoid robots. For example, when pressure is detected by the tactile sensor at the wrist, the corresponding IK target is shifted slightly along the surface normal in proportion to the measured pressure, helping to mitigate internal forces during physical interaction\footnote{Although the admittance control was verified to function as intended on the real humanoid robot, it was applied during manipulation experiments only in simulation in this study due to constraints of the robot setup.}.


\section{Learning-based Manipulation Control} \label{sec:manip-learn}

\subsection{Model Structure}

The challenge of managing distributed tactile measurements in control stems from their high dimensionality. For instance, in the distributed tactile sensors mounted on the upper body of the humanoid in this study, the intensity of touch and proximity is measured for each of the 163 cells. To extract skills from such high-dimensional data, we apply Transformer-based imitation learning. Specifically, we extend ACT~\cite{ALOHA:Zhao:RSS2023}, a highly effective method in bimanual manipulation, to enable the handling of tactile modality. In this study, we use the term ``tactile modality'' to refer not only to direct touch sensing but also to proximity sensing, which captures near-contact interactions.


\begin{figure}[tpb]
  \centering
  \includegraphics[width=0.98\columnwidth]{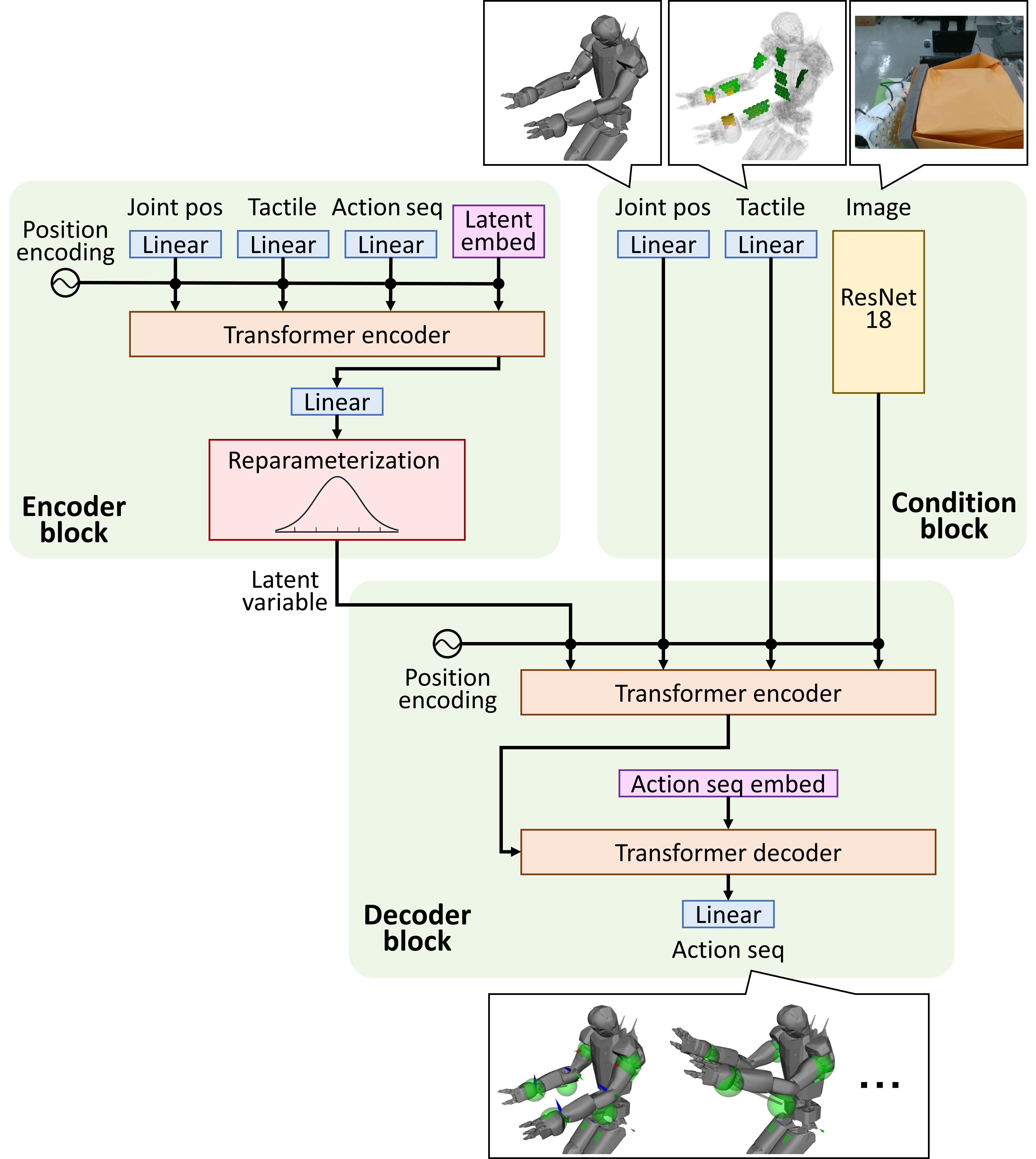}
  \caption{Model structure of TACT (tactile-modality extended ACT).
    \newline
    \footnotesize{TACT is a CVAE-based policy that takes as input joint positions, images, and tactile measurements. The encoder encodes sensory inputs and actions, the condition block extracts features from current observations, and the decoder generates temporally consistent action sequences using Transformers. The figure visualizes examples of input and output data to the model in humanoid manipulation experiments.}}
  \label{fig:act}
\end{figure}

\figref{fig:act} illustrates the model structure of TACT, which extends ACT by incorporating tactile sensing. Like ACT, the model is based on a conditional variational autoencoder (CVAE)~\cite{CVAE:Sohn:NeurlPS2015}, and consists of an encoder, a decoder, and a condition block. The encoder and decoder are implemented using Transformers~\cite{Transformer:Vaswani:NeurlPS2017}, while ResNet18~\cite{ResNet:He:CVPR2016} is used for image feature extraction. The key difference from ACT lies in the inclusion of tactile measurements: they are flattened into a 1D vector, linearly projected into a token, and fed into both the encoder and condition blocks. This allows the Transformer to learn correlations between tactile input and other modalities for whole-body contact manipulation.


\subsection{Model Training and Deployment}

The TACT policy is trained using data obtained by a human operating a humanoid robot to perform whole-body contact manipulation through the retargeting control described in Section~\ref{sec:retarget-locomotion}. The training data consists of pairs of arm joint positions, tactile measurements, color images captured by a head camera, and actions.
It should be noted that the actions are not the robot's joint positions $\bm{q}^{\mathrm{cmd}}$, but rather the poses $\bm{\hat{p}}_{\mathrm{arm}}^{\mathrm{cmd}}$ of the human arms. This design maintains consistency between training and deployment by allowing the same input-output structure, and avoids learning joint-space feedback behavior such as compliance, which is handled separately by the lower-layer controller.

The policy is trained as a standard CVAE by minimizing the KL divergence of latent variables and the error of predicted actions. During deployment, the encoder block is omitted, and the decoder generates a sequence of future actions from sensor inputs provided to the condition block, using a zero latent vector. Thanks to the LIPM-based biped controller, the robot can maintain balance and continue walking even when its upper body is governed by the learned policy.



\section{Experiments} \label{sec:exp}

\subsection{Implementation}

\subsubsection{Humanoid with Tactile Sensors}

The distributed tactile sensor e-skin from Intouch Robotics~\cite{Cheng:RobotSkin:IEEE2019} was mounted on the upper body (chest, upper arms, forearms, and wrists) of the RHP7 Kaleido, a life-size humanoid robot developed by Kawasaki Heavy Industry~\cite{Kakiuchi:RHP:IROS2017}, as shown in \figref{fig:intro}. The e-skin is a sheet-shaped sensor with hexagonal cells arranged in two dimensions. The sensor is about 5~mm thick and can be deformed so that it can be arranged along the body surface of the robot. We divided the sensors into 11 patches for each body link of the robot, and a total of 163 cells were mounted on the robot. Each cell can measure the intensity of tactile and proximity sensations, as well as acceleration. The effectiveness of our method is ensured by the reproducibility of the sensor measurements, obviating the necessity for calibration to convert the measured intensities into physical quantities such as force or distance.


\subsubsection{Humanoid Teleoperation System}

To collect learning data, we constructed a teleoperation system that measures the posture of a human operator and retargets it to a humanoid. We employed the trackers and hand controllers of the HTC VIVE to measure the posture of the human operator. The operator was equipped with trackers on their waist and both elbows, and held the hand controllers in both hands to measure the poses of these parts. The operator was able to control the opening and closing of the robot's grippers in response to the trigger buttons on the hand controllers. When performing manipulation by teleoperation, the robot and operator were positioned facing each other as shown in \figref{fig:intro} to allow the operator to directly see the robot's workspace. In order to facilitate intuitive teleoperation, a left-right inversion was applied in retargeting, thereby mapping the operator's right arm to the robot's left arm and vice versa.


\subsubsection{Software Framework}

The retargeting and locomotion control was implemented within the real-time robot control framework {\tt mc\_rtc}~\cite{mc_rtc:github2025}. In this system, kinematic commands such as the target poses of the CoM, feet, and arms are passed to an acceleration-based whole-body IK calculation. The resulting joint positions $\bm{q}^{\mathrm{cmd}}$ are sent to a low-level joint PD controller. In addition to the distributed tactile sensors, the controller uses the joint positions from the joint encoders, the contact wrench from the 6-axis force/torque sensors mounted on the feet, and the body orientation from the IMU sensor mounted on the waist link. Tactile sensor measurements are flattened into a 1D vector of 326 elements, representing the intensity (non-negative scalar values) of tactile and proximity sensing across 163 sensor cells, and are input to the linear layer of the manipulation policy. As in the original ACT~\cite{ALOHA:Zhao:RSS2023}, TACT applied a temporal ensemble that takes the weighted average of the predicted action history with a chunk size of 20 (corresponding to 2~s). The retargeting and locomotion control runs at 500~Hz, while the manipulation policy operates at 10~Hz.


\subsection{Real-world Experiments} \label{sec:exp-real}

We evaluated the success rate of the same task with the proposed {\bf TACT} policy and three baseline policies. The first baseline is a policy that replays the human teleoperation data as-is ({\bf Replay}). The other two baselines are policies that remove the vision modality from TACT ({\bf TACT w/o vision}) and the tactile modality ({\bf TACT w/o tactile}, which is identical to the original ACT).


Since the tactile modality is particularly important for manipulations that require fine adjustment of the contact force, we chose a task in which the robot uses both arms to hold up a fragile box made of thin paper folded and glued with sponge on both sides. To prevent the object from rotating, the robot needs to make contact with the surface of the forearm and wrist, which requires richer contact than pick-and-place with a gripper. In addition, because paper boxes are fragile and easily crushed, the delicate contact must be controlled to avoid applying too much force.


In the training phase, we prepared a dataset consisting of 27 episodes in which the robot held up three different sized boxes (440~mm, 270~mm, and 210~mm in width, respectively) for 9 times each while being teleoperated by a human operator. \figref{fig:act} shows an example of the visualized data. In the deployment phase, we verified whether the robot could autonomously hold up a box by applying the proposed and baseline policies trained on this dataset. A single learned policy is deployed across boxes of varying sizes, without explicit size information provided at test time. The policy must instead infer object size implicitly from visual and tactile observations. In both the training and deployment phases, the boxes were randomly placed on the table in front of the robot. In order to evaluate the ability to perform delicate whole-body contact manipulation, we judged the task as failed not only when the box was dropped, but also when the box was crushed to the extent that its shape was distorted\footnote{In order to ensure fair evaluation across policies, the following three points were considered: (1) the failure judgment due to crushing of the box was made by a human who did not know which policy was being applied, (2) to mitigate the effect of the box changing state during the repetition of trials, the four policies were switched one at a time and executed in the order of the cycle, (3) every policy was executed once for the box placed at the same randomly chosen position.}.

\begin{figure}[tpb]
  \centering
  \includegraphics[width=0.24\columnwidth, trim=0 9cm 0 0, clip]{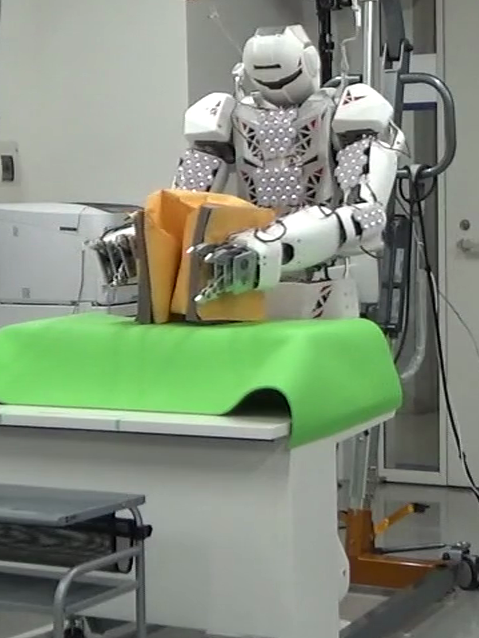}
  \includegraphics[width=0.24\columnwidth, trim=0 9cm 0 0, clip]{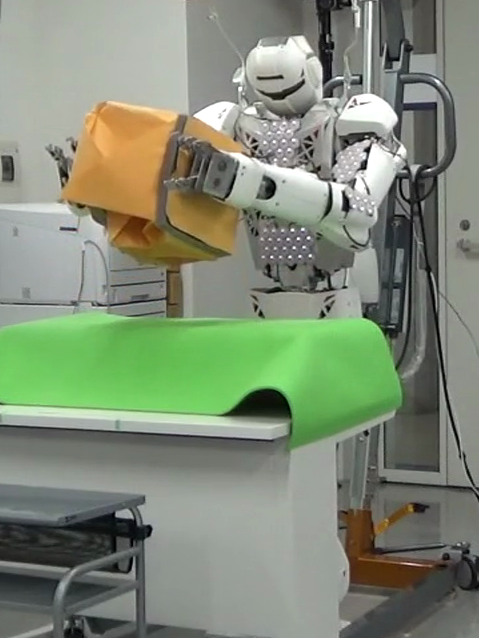}
  \includegraphics[width=0.24\columnwidth, trim=0 9cm 0 0, clip]{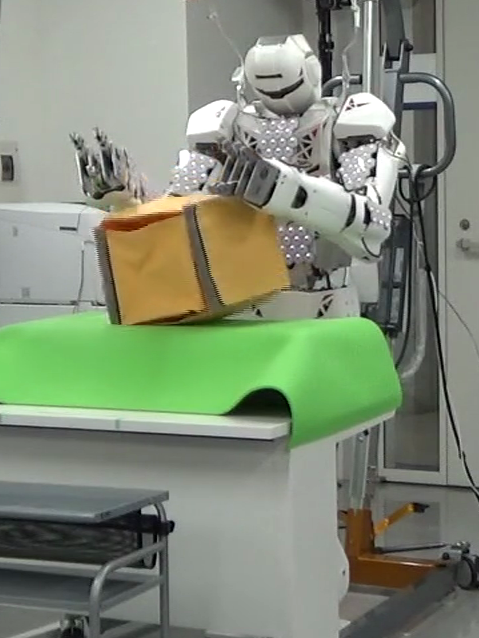}
  \includegraphics[width=0.24\columnwidth, trim=0 9cm 0 0, clip]{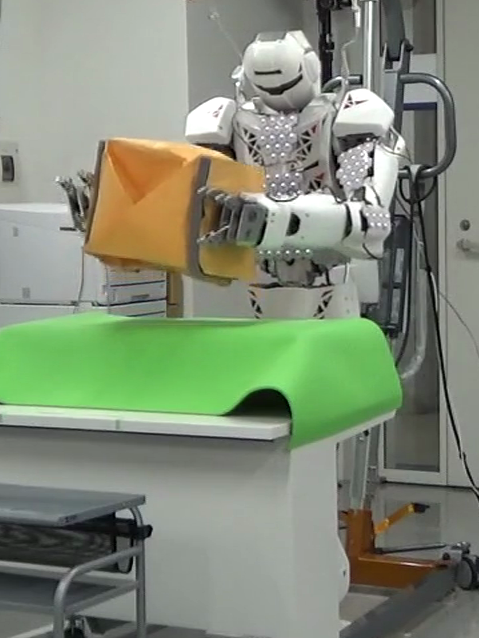}\\
  \vspace{-1.2mm}
  \begin{minipage}{0.24\columnwidth}
    \begin{center} \scriptsize Replay \end{center}
  \end{minipage}
  \begin{minipage}{0.24\columnwidth}
    \begin{center} \scriptsize TACT w/o vision \end{center}
  \end{minipage}
  \begin{minipage}{0.24\columnwidth}
    \begin{center} \scriptsize TACT w/o tactile \end{center}
  \end{minipage}
  \begin{minipage}{0.24\columnwidth}
    \begin{center} \scriptsize TACT (proposed) \end{center}
  \end{minipage}\\
  \vspace{2mm}
  \footnotesize (A) Manipulation of a medium-size box\\
  \vspace{3mm}
  \includegraphics[width=0.24\columnwidth, trim=0 9cm 0 0, clip]{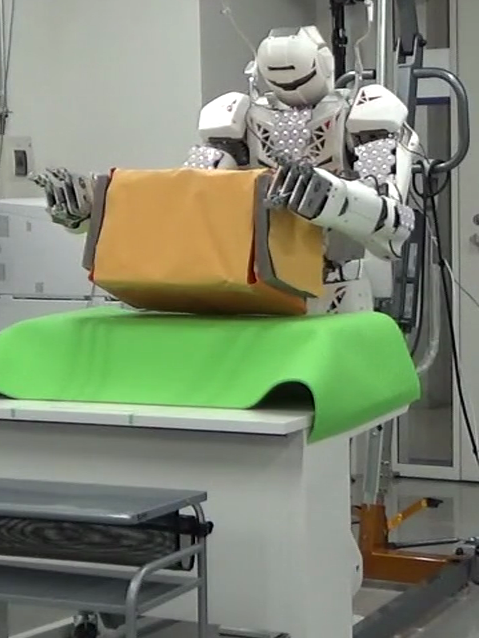}
  \includegraphics[width=0.24\columnwidth, trim=0 9cm 0 0, clip]{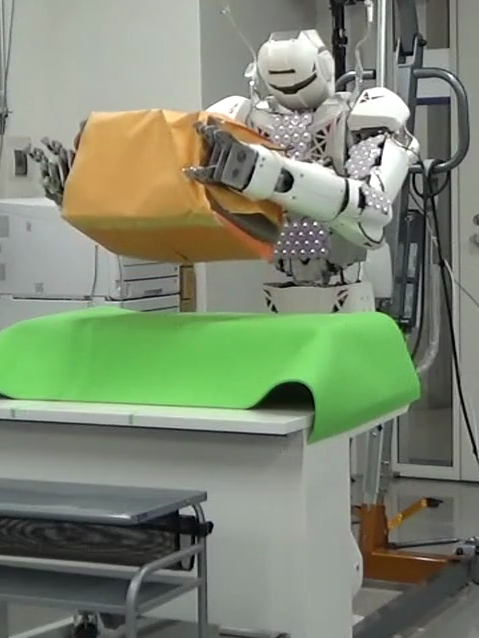}
  \includegraphics[width=0.24\columnwidth, trim=0 9cm 0 0, clip]{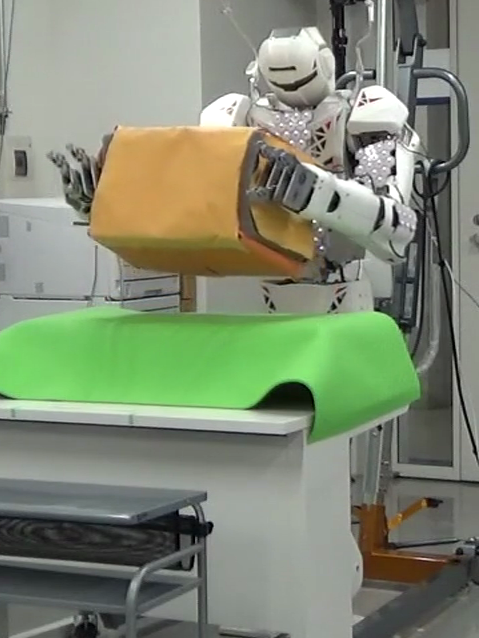}
  \includegraphics[width=0.24\columnwidth, trim=0 9cm 0 0, clip]{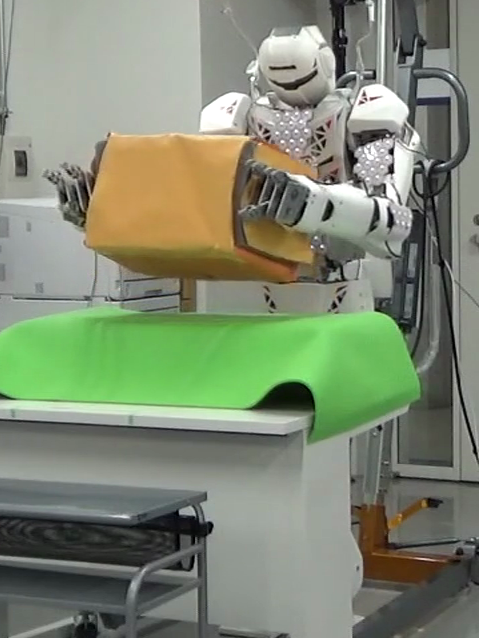}\\
  \begin{minipage}{0.24\columnwidth}
    \begin{center} \scriptsize Replay \end{center}
  \end{minipage}
  \begin{minipage}{0.24\columnwidth}
    \begin{center} \scriptsize TACT w/o vision \end{center}
  \end{minipage}
  \begin{minipage}{0.24\columnwidth}
    \begin{center} \scriptsize TACT w/o tactile \end{center}
  \end{minipage}
  \begin{minipage}{0.24\columnwidth}
    \begin{center} \scriptsize TACT (proposed) \end{center}
  \end{minipage}\\
  \vspace{2mm}
  \footnotesize (B) Manipulation of a large-size box\\
  \vspace{3mm}
  \includegraphics[width=0.24\columnwidth, trim=0 9cm 0 0, clip]{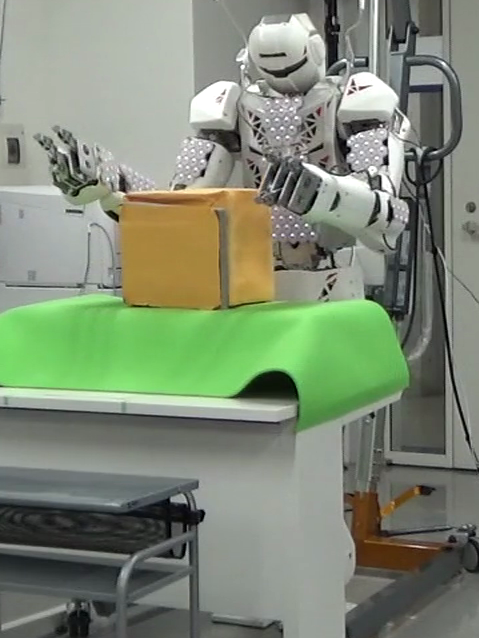}
  \includegraphics[width=0.24\columnwidth, trim=0 9cm 0 0, clip]{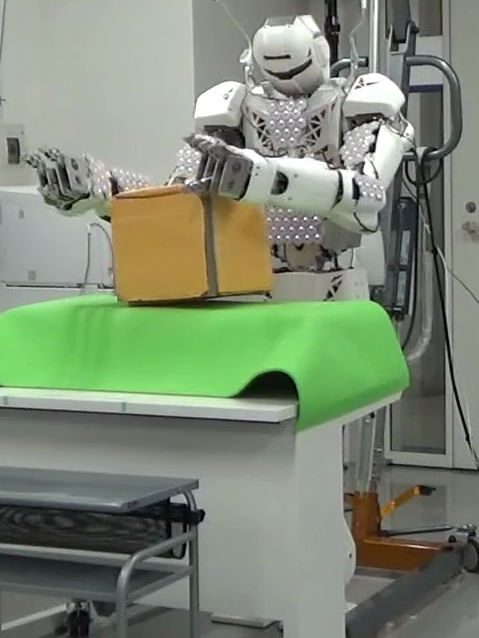}
  \includegraphics[width=0.24\columnwidth, trim=0 9cm 0 0, clip]{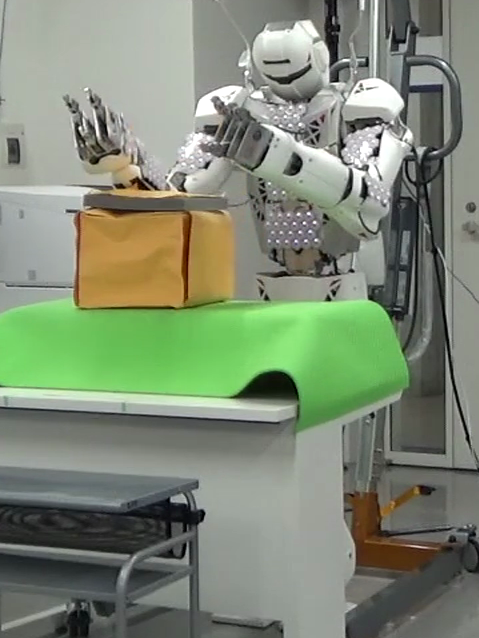}
  \includegraphics[width=0.24\columnwidth, trim=0 9cm 0 0, clip]{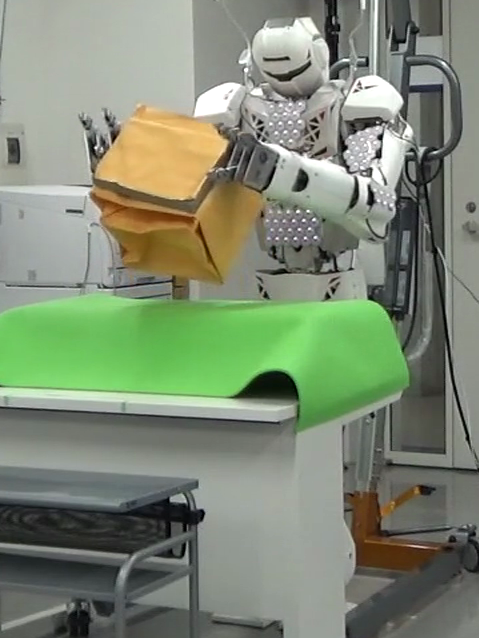}\\
  \begin{minipage}{0.24\columnwidth}
    \begin{center} \scriptsize Replay \end{center}
  \end{minipage}
  \begin{minipage}{0.24\columnwidth}
    \begin{center} \scriptsize TACT w/o vision \end{center}
  \end{minipage}
  \begin{minipage}{0.24\columnwidth}
    \begin{center} \scriptsize TACT w/o tactile \end{center}
  \end{minipage}
  \begin{minipage}{0.24\columnwidth}
    \begin{center} \scriptsize TACT (proposed) \end{center}
  \end{minipage}\\
  \vspace{2mm}
  \footnotesize (C) Manipulation of an unseen-size box\\
  \caption{Experiment in which a humanoid holds up a paper box.
    \newline
    \footnotesize{
(A) For the medium-size box, Replay and TACT w/o vision failed by crushing the box, and TACT w/o tactile failed by dropping it, whereas TACT succeeded by gently lifting the box.
(B) For the large-size box, Replay dropped the box and TACT w/o vision crushed it, but both TACT w/o tactile and TACT succeeded.
(C) For the unseen-size box, the robot attempted to lift the medium-size box by contacting the side without a sponge. This object configuration was not included in the training data. While all baseline policies failed, only TACT succeeded.
  }}
  \label{fig:exp-box}
\end{figure}

\figref{fig:exp-box} shows the deployment of the policies, and Table~\ref{tab:exp-box} shows the evaluation results. TACT (the proposed method) succeeded in lifting the medium-size and the large-size boxes in all trials, while the baseline policies had failures and a low success rate. TACT w/o vision tended to crush the box, and TACT w/o tactile tended to drop the medium-size box when lifting it because of a small gap between the arm and the box. In Replay, demonstration data was randomly selected and played in feedforward regardless of box size or position, so failures such as dropping or crushing occurred randomly. TACT also showed superiority for unseen-size boxes between large and small sizes that were not included in the training data. In the small-size boxes, the robot succeeded in reaching the arms into the box in all policies, but failed to maintain contact when lifting it, causing the box to fall. This outcome is presumably attributable to the substantial variation in box sizes present in the training data. Specifically, the widths of the small- and large-size boxes differed by more than a factor of two. Consequently, the generalization accuracy of the policies for box size was inadequate. We expect that this problem can be mitigated by increasing the number of episodes and enriching the variation of box sizes in the training data. Overall, these results confirm that policies that handle both visual and tactile modalities, such as TACT, are essential for delicate whole-body contact manipulation. We believe this is because vision is necessary to discriminate between the boxes with significantly different sizes, and tactile sensation is necessary for delicate control to maintain contact between the robot arm and the box.


\renewcommand{\arraystretch}{1.6}
\begin{table}[tpb]
  \caption{Success rates for holding up a box of different sizes.}
  \label{tab:exp-box}
  \vspace{-3mm}
  \begin{center}
    \begin{tabular}{l||cccc|c}
      \Xhline{1pt}
      & Small & Medium & Large & Unseen & Total\\
      \hline
      Replay           & \bm{2 / 3} & 0 / 3      & 0 / 3      & 1 / 5      & 3 / 14      \\
      TACT w/o vision  & 0 / 3      & 0 / 3      & 1 / 3      & 3 / 5      & 4 / 14      \\
      TACT w/o tactile & 0 / 3      & 1 / 3      & \bm{3 / 3} & 2 / 5      & 6 / 14      \\
      TACT (proposed)  & 0 / 3      & \bm{3 / 3} & \bm{3 / 3} & \bm{4 / 5} & \bm{10 / 14}\\
      \Xhline{1pt}
    \end{tabular}
  \end{center}
\end{table}

\begin{figure}[tpb]
  \begin{minipage}{0.48\linewidth}
    \centering
    \includegraphics[width=0.775\columnwidth]{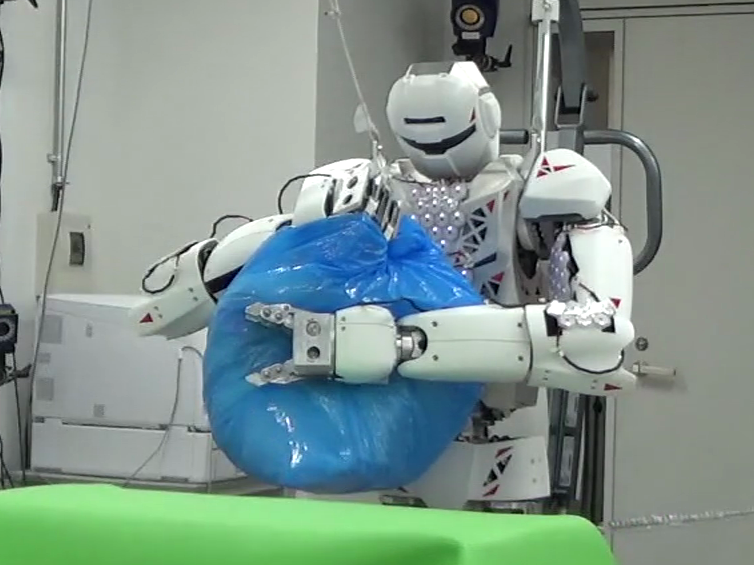}\\
    \vspace{1mm}
    \includegraphics[width=0.775\columnwidth]{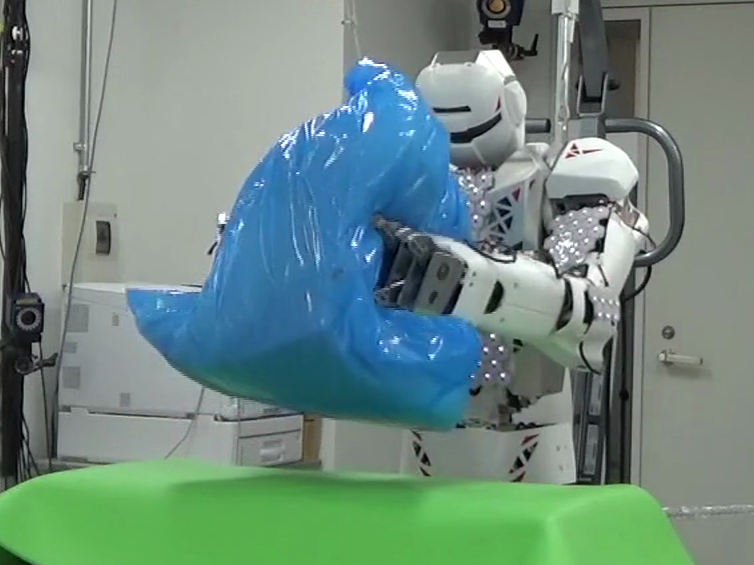}
    \caption{Holding up a large plastic bag autonomously.}
    \label{fig:exp-bag}
  \end{minipage}
  \hfill
  \begin{minipage}{0.48\linewidth}
    \centering
    \includegraphics[width=0.885\columnwidth]{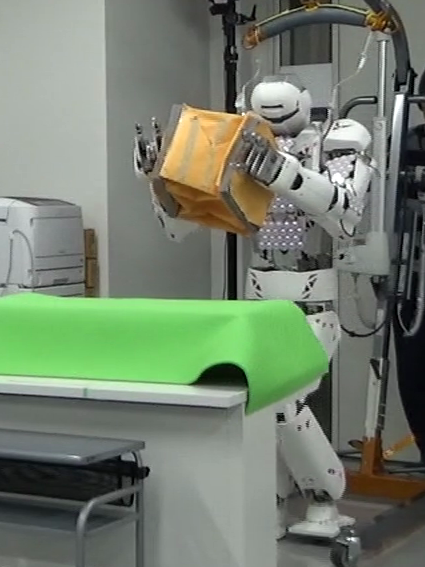}
    \caption{Bipedal walking while holding up an object.}
    \label{fig:exp-walk}
  \end{minipage}
\end{figure}

We also confirmed that the proposed method can be applied to other objects, as shown in \figref{fig:exp-bag}. In this experiment, the robot lifts a large plastic bag in two different ways; one is to grasp the top of the bag with one hand and hold it with the other arm, and the other is to hold it with both arms from both sides. For each motion, we collected 8 episodes of demonstration data and trained individual policies. It was shown that the proposed method can be applied to manipulation motions in which the robot contacts the object simultaneously at many parts of the body, including the forearm and the chest.

\begin{figure}[tpb]
  \centering
  \vspace{-1.5mm}
  \begin{minipage}[c]{0.48\columnwidth}
    \centering
    \includegraphics[width=0.82\columnwidth]{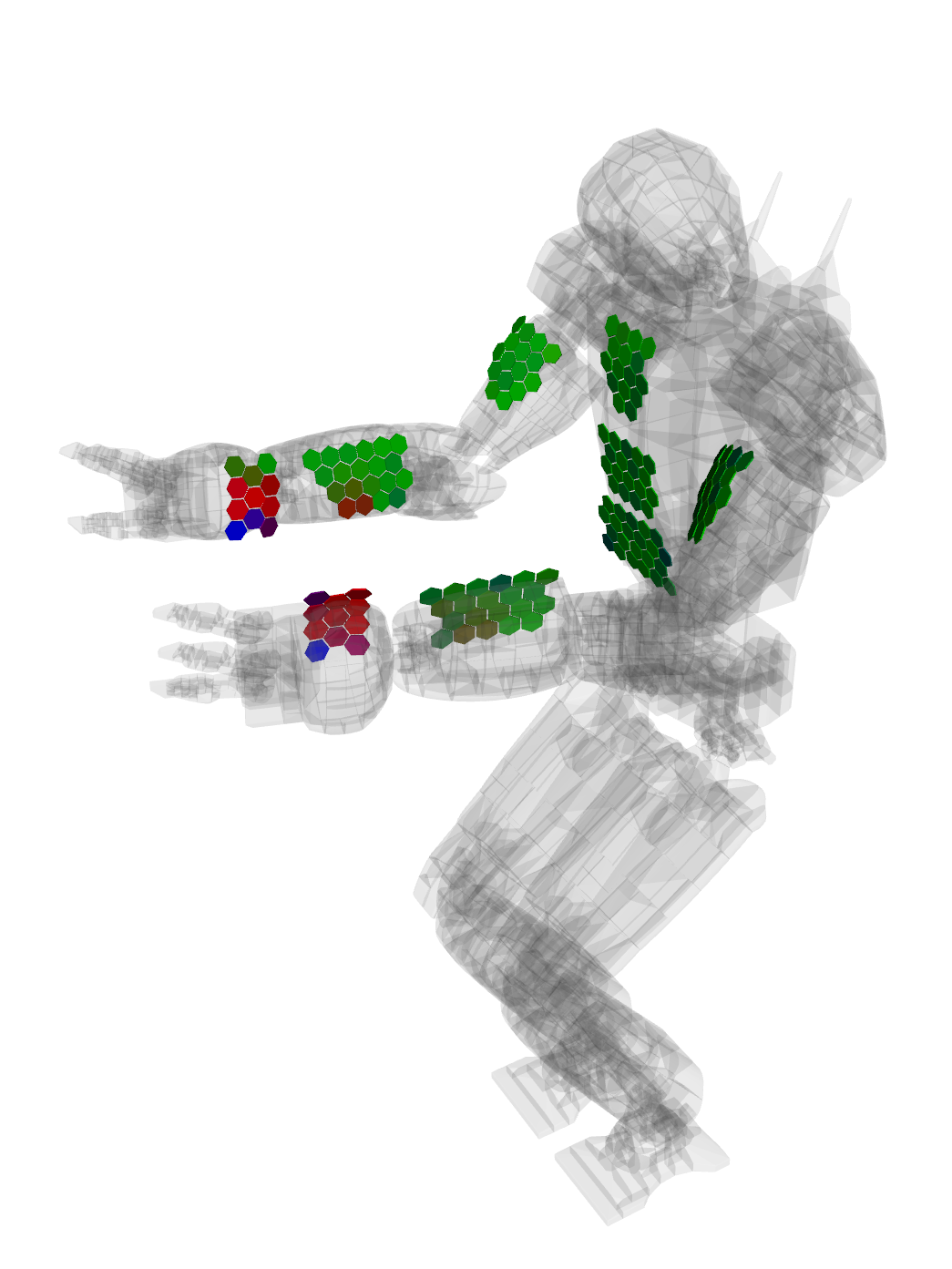}
  \end{minipage}
  \begin{minipage}[c]{0.48\columnwidth}
    \centering
    \includegraphics[width=0.8\columnwidth]{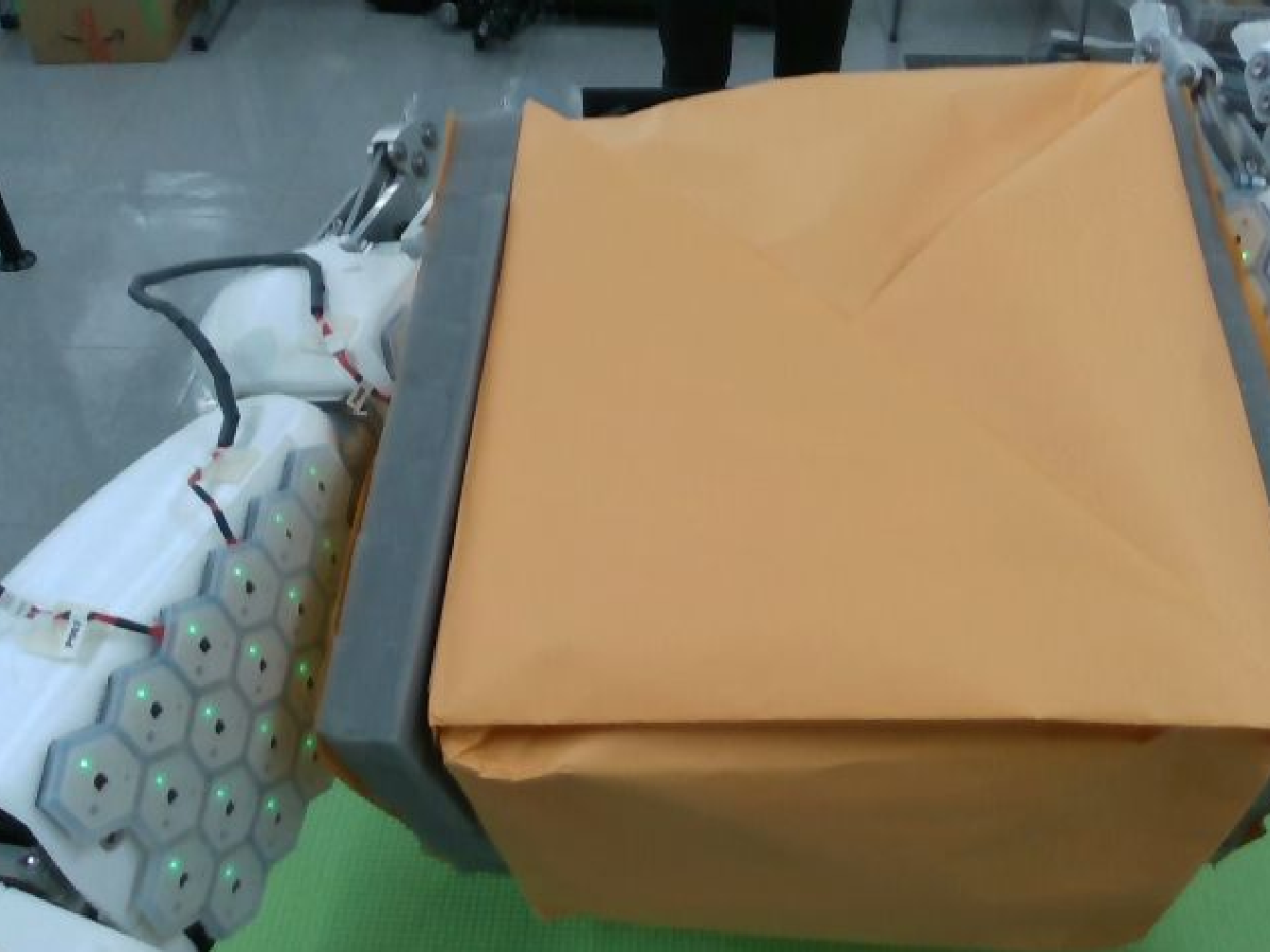}
  \end{minipage}\\
  \begin{minipage}{0.48\columnwidth}
    \begin{center} \footnotesize (A) Tactile measurements \end{center}
  \end{minipage}
  \begin{minipage}{0.48\columnwidth}
    \begin{center} \footnotesize (B) Head camera image \end{center}
  \end{minipage}\\
  \vspace{3mm}
  \begin{minipage}[c]{0.49\columnwidth}
    \centering
    \includegraphics[width=1.0\columnwidth]{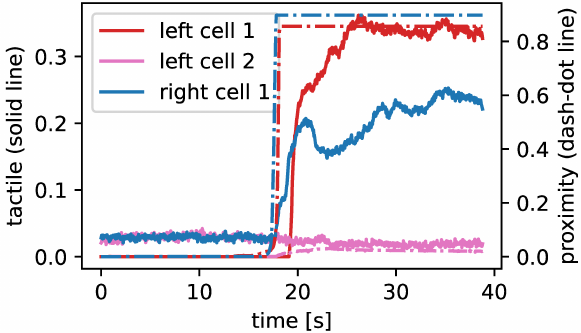}
  \end{minipage}
  \begin{minipage}[c]{0.49\columnwidth}
    \centering
    \includegraphics[width=1.0\columnwidth]{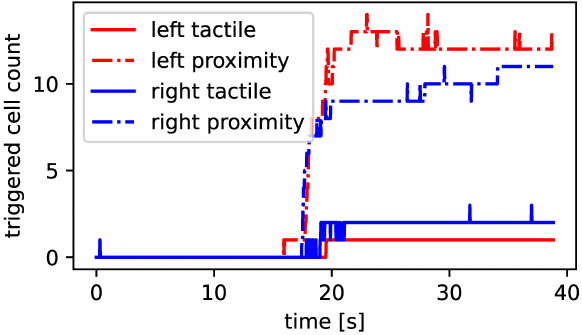}
  \end{minipage}\\
  \vspace{1mm}
  \begin{minipage}{0.48\columnwidth}
    \begin{center} \footnotesize (C) Tactile and proximity signals \end{center}
  \end{minipage}
  \begin{minipage}{0.48\columnwidth}
    \begin{center} \footnotesize (D) Active sensor cell count \end{center}
  \end{minipage}\\
  \vspace{3mm}
  \begin{minipage}[c]{0.325\columnwidth}
    \centering
    \includegraphics[width=1.0\columnwidth]{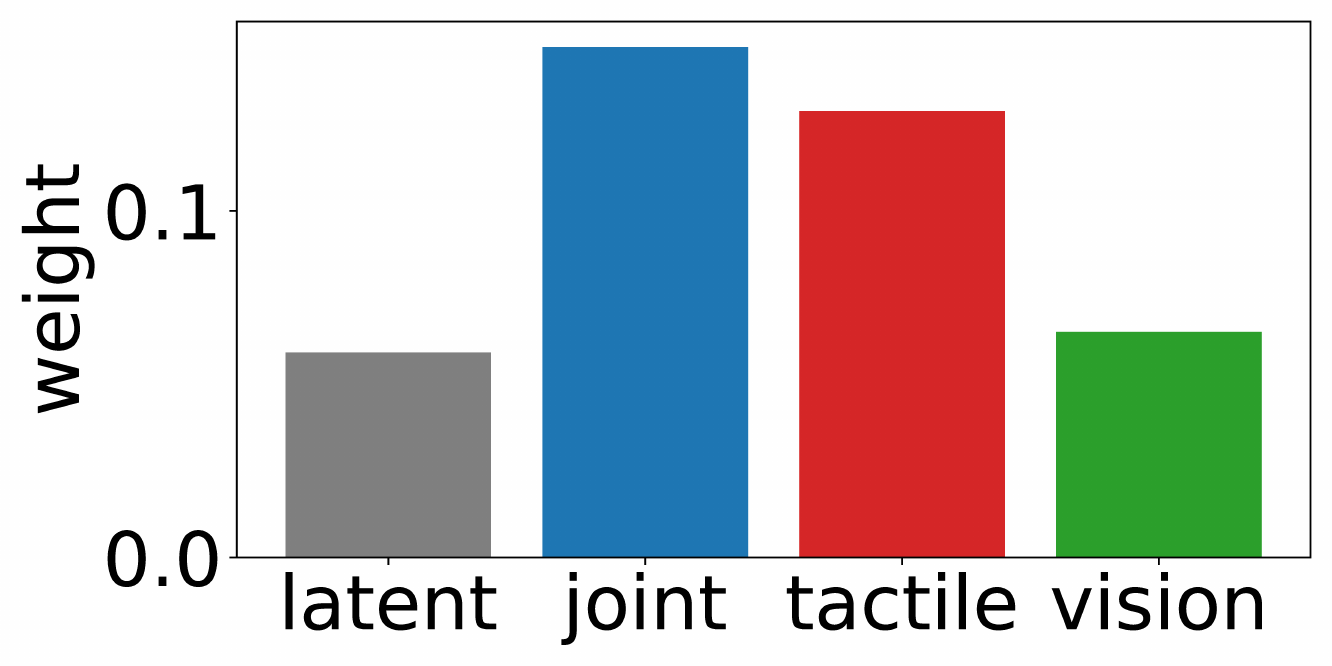}
  \end{minipage}
  \begin{minipage}[c]{0.325\columnwidth}
    \centering
    \includegraphics[width=1.0\columnwidth]{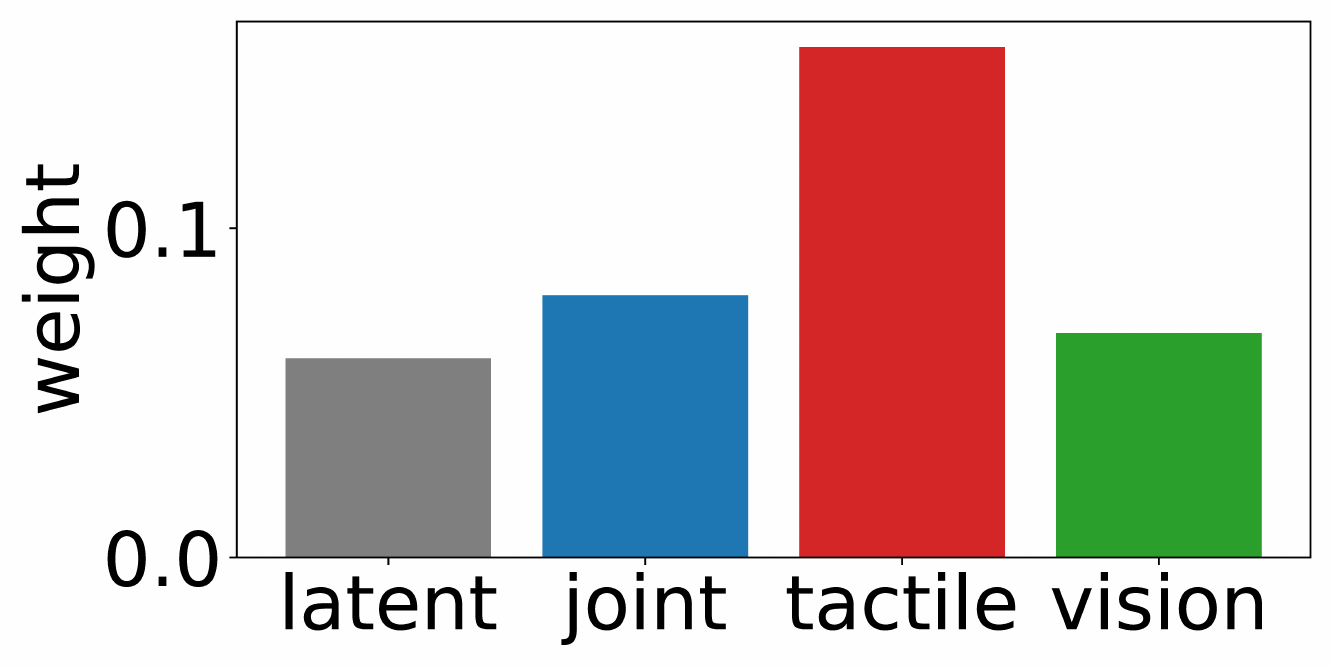}
  \end{minipage}
  \begin{minipage}[c]{0.325\columnwidth}
    \centering
    \includegraphics[width=1.0\columnwidth]{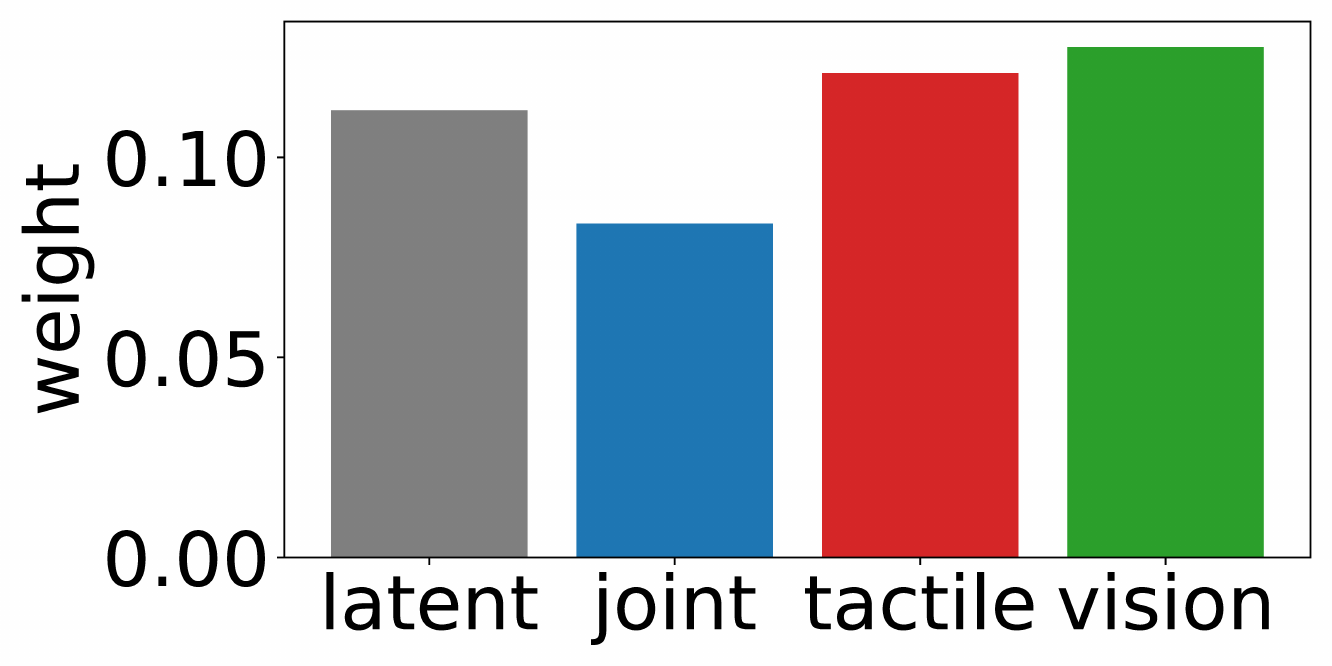}
  \end{minipage}\\
  \begin{minipage}{0.325\columnwidth}
    \begin{center} \scriptsize \hspace{4mm} t = 5 s \end{center}
  \end{minipage}
  \begin{minipage}{0.325\columnwidth}
    \begin{center} \scriptsize \hspace{4mm} t = 15 s \end{center}
  \end{minipage}
  \begin{minipage}{0.325\columnwidth}
    \begin{center} \scriptsize \hspace{4mm} t = 30 s \end{center}
  \end{minipage}\\
  \begin{minipage}{1.0\columnwidth}
    \begin{center} \footnotesize (E) Transformer attention weights \end{center}
  \end{minipage}\\
  \caption{Sensor and policy data during paper box manipulation.
    \newline
    \footnotesize{
      (A)~Proximity and tactile signal magnitudes are visualized in red and blue, respectively. Strong proximity signals are detected in the cells on the left and right wrists and forearms, indicating the object is held between the arms. Tactile signals are also observed in some wrist cells.
      (B)~The left arm is visible in the head camera view, while the right arm is mostly out of view.
      (C)~Time series of representative tactile and proximity signals recorded during policy deployment. Three sensor cells are sampled for plotting: ``left cell 1'' and ``right cell 1'' experienced contact during the task, while ``left cell 2'' did not.
      (D)~Number of sensor cells activated by tactile or proximity signals during policy deployment. For visualization, cells on the left and right arms are counted separately, with a threshold of 0.1 for activation.
      (E)~Attention weights from the Transformer at three time points corresponding to the horizontal axis in plots (C) and (D). For each frame, we extract a column of the attention matrix (i.e., $\mathrm{softmax}(\bm{Q} \bm{K}^{\mathsf{T}} / \sqrt{d_k})$ in~\cite{Transformer:Vaswani:NeurlPS2017}) corresponding to the joint position token, which is directly related to action prediction. For vision input, we report the average attention weight across all 12 image tokens. Strong attention weights appear on the tactile input shortly before contact events, suggesting that tactile input is actively utilized by the policy.
  }}
  \label{fig:exp-data-box}
  \vspace{3mm}
  \begin{minipage}[c]{0.48\columnwidth}
    \centering
    \includegraphics[width=0.73\columnwidth]{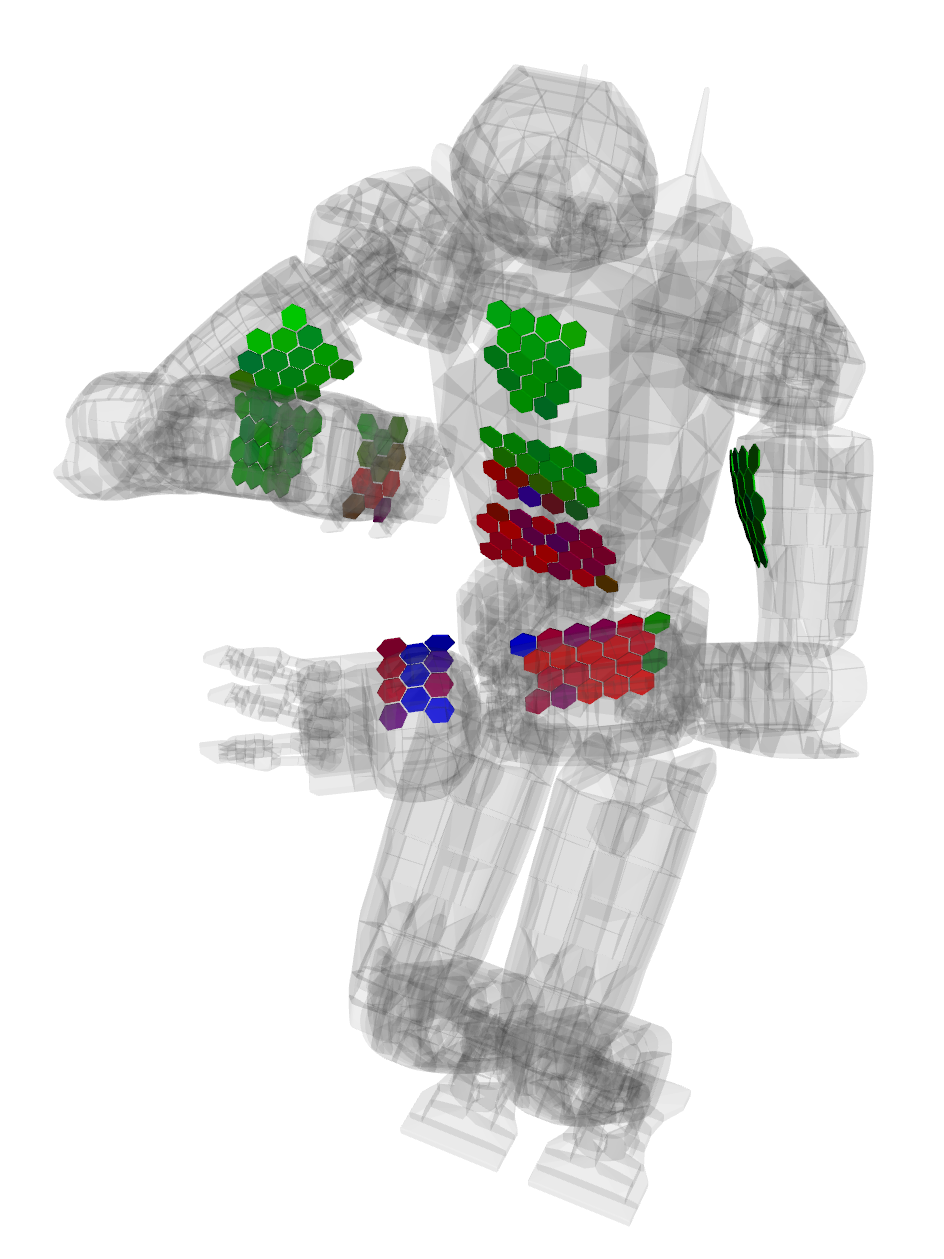}
  \end{minipage}
  \begin{minipage}[c]{0.48\columnwidth}
    \centering
    \includegraphics[width=0.8\columnwidth]{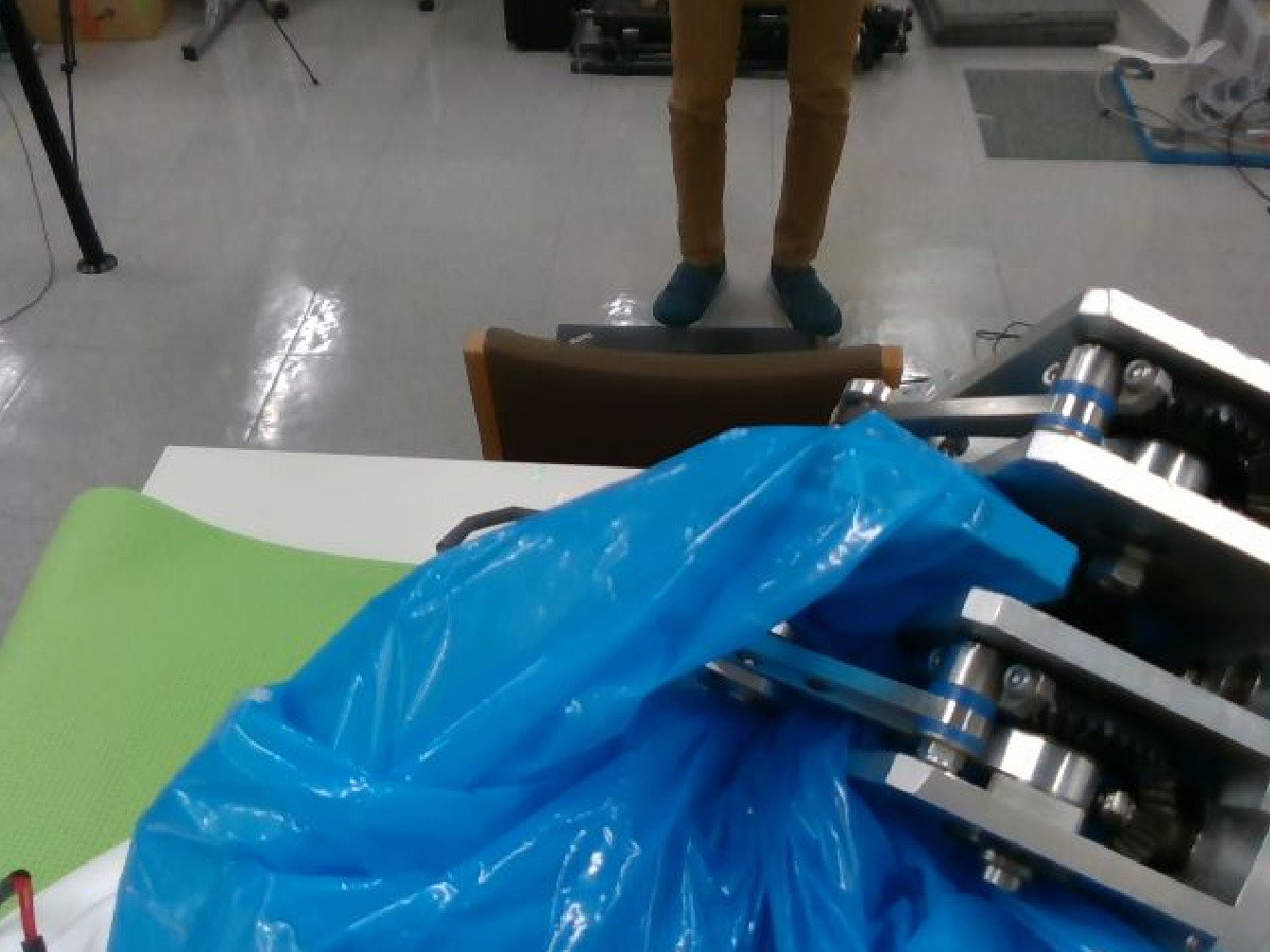}
  \end{minipage}\\
  \begin{minipage}{0.48\columnwidth}
    \begin{center} \footnotesize (A) Tactile measurements \end{center}
  \end{minipage}
  \begin{minipage}{0.48\columnwidth}
    \begin{center} \footnotesize (B) Head camera image \end{center}
  \end{minipage}
  \caption{Sensor data during plastic bag manipulation.
    \newline
    \footnotesize{
      (A)~Proximity and tactile signal magnitudes are visualized in red and blue, respectively. Strong proximity signals are detected in the cells of the chest, right wrist, and left forearm and wrist, indicating the object is enclosed by the arms and chest. Tactile signals in the cells of the chest and left wrist suggest that the object is pressed between the chest and arm.
      (B)~Due to close contact with the chest, much of the object and arms are out of view or occluded in the head camera image.
  }}
  \label{fig:exp-data-bag}
\end{figure}

Figs.~\ref{fig:exp-data-box} and \ref{fig:exp-data-bag} show examples of sensor measurements during the manipulation of a paper box and a plastic bag, respectively. During these tasks, the short distance between the robot and the object makes it difficult for the head-mounted camera to capture both the arms and the object in full view. Nevertheless, tactile measurements provide rich information about the location and intensity of contact, enabling the policy to operate effectively even under such conditions.
To further illustrate the role of tactile sensing, \figref{fig:exp-data-box}~(C)-(E) presents time-series plots of representative tactile and proximity signals, the number of activated sensor cells, and Transformer attention weights. These visualizations highlight strong attention to tactile input around contact events, suggesting that the policy leverages tactile information during manipulation.


In all experiments reported in this paper, the robot successfully performed manipulation tasks while standing on two legs without losing balance, both during teleoperation and policy inference. Furthermore, as shown in \figref{fig:exp-walk}, the robot was able to walk stably while carrying an object in both phases. These results demonstrate the effectiveness of the proposed system, which combines model-based retargeting and locomotion control with learning-based manipulation control, in achieving reliable and flexible loco-manipulation with a life-size humanoid robot.


\subsection{Simulation Experiments}

To further evaluate the proposed method, we constructed a simulation environment using the MuJoCo physics engine. The robot was equipped with simulated tactile sensor patches composed of hexagonally arranged cells, replicating the configuration of the real robot. For simplicity, proximity sensing was not included in the simulation.

\begin{figure}[tpb]
  \centering
  \includegraphics[width=1.0\columnwidth]{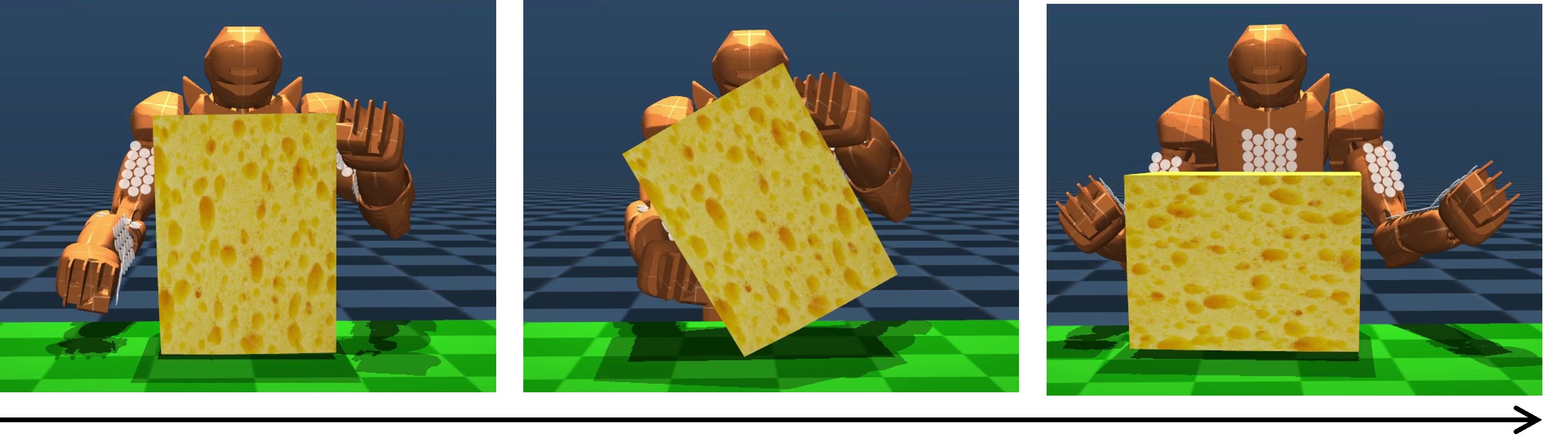}\\
  \begin{minipage}{1.0\columnwidth}
    \begin{center} \footnotesize (A) Box reorientation task using both arms \end{center}
  \end{minipage}\\
  \vspace{2mm}
  \includegraphics[width=1.0\columnwidth]{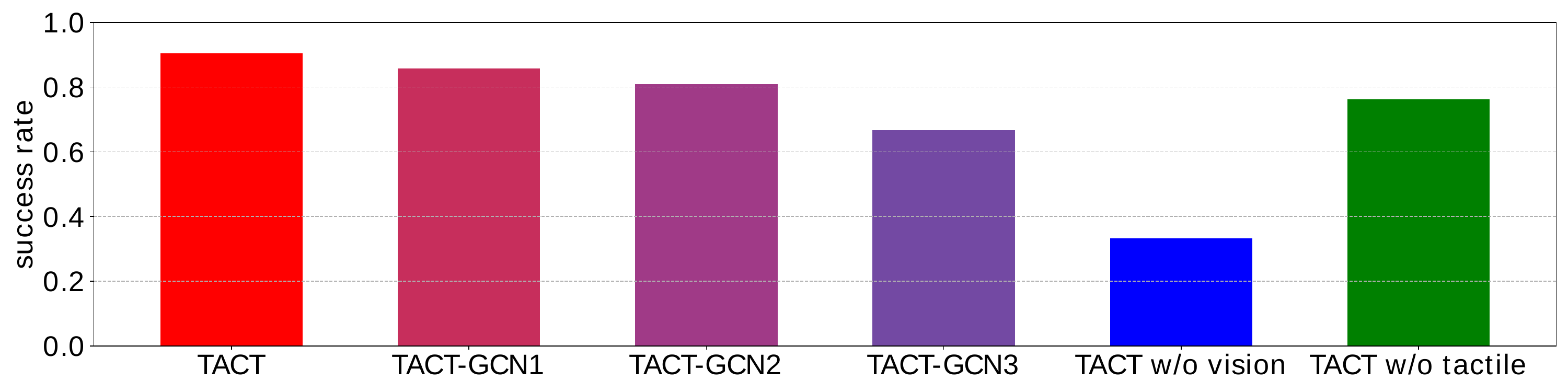}\\
  \begin{minipage}{1.0\columnwidth}
    \begin{center} \footnotesize (B) Success rates of each policy \end{center}
  \end{minipage}
  \caption{Simulation task and results.
    \newline
    \footnotesize{
      (B) TACT-GCN1, TACT-GCN2, and TACT-GCN3 denote variants of TACT with 1, 2, and 3 graph convolutional layers, respectively.
  }}
  \label{fig:sim-main}
\end{figure}

The robot performed a box reorientation task with complex contact transitions, as shown in \figref{fig:sim-main}~(A), tilting a box on a table by 90 degrees using both arms. We collected 30 teleoperated demonstrations with varying initial box positions, using the same interface as in the real-world experiments.

Each trained policy was deployed 21 times with different box positions, and the resulting success rates are shown in \figref{fig:sim-main}~(B). The proposed TACT policy outperformed the baselines TACT w/o vision and TACT w/o tactile, showing consistent results with the real-world experiments despite differences in domain and task.

We also evaluated a TACT variant that adds a graph convolutional network (GCN) before the linear layer for tactile feature extraction. The GCN used an adjacency matrix based on the geometric connectivity of sensor cells. This structure was intended to leverage spatial relationships among sensor cells. However, TACT-GCN slightly underperformed compared to TACT, suggesting that in this setting, a simple linear projection is already effective. Incorporating spatial structure remains a non-trivial challenge for future work.

\section{Conclusion}

In this paper, we proposed a humanoid control system with deep imitation learning to robustly perform whole-body contact manipulation with delicate contact. We introduced a new model called TACT, which adds the tactile modality to ACT, the existing major imitation learning model. To collect data for training the model, we developed an intuitive teleoperation system that allows retargeting from a human to a humanoid. By integrating a learning-based manipulation control with model-based retargeting and locomotion control, we have demonstrated that the RHP7 Kaleido, a life-size humanoid robot mounted with tactile sensors on its upper body, can reliably and flexibly perform contact-rich whole-body manipulation. Experiments have shown that both visual and tactile modalities are essential for successful manipulation involving broad and delicate contact.
This is the first study to achieve whole-body contact loco-manipulation by a life-size humanoid robot based on imitation learning with multimodal sensor inputs involving visual and tactile information.


Future work includes replacing the model-based retargeting and locomotion controller with a reinforcement learning policy that accepts retargeting commands as input~\cite{OmniH2O:He:CoRL2024,HumanPlus:Fu:CoRL2024}. Since the proposed tactile-based manipulation policy is independent of the lower-layer controller, this integration could enable more robust loco-manipulation while preserving complex whole-body contact skills.
Another direction is to extend the method to stiff or heavy objects, which would broaden its applicability. However, with position-controlled robots like the one used in this study, internal forces arising from rigid contact can induce significant joint load. This suggests that hardware with higher backdrivability may be needed. Additionally, the locomotion controller must be enhanced to maintain balance under heavier loads.
Finally, further exploration of policy architectures is a promising direction. This includes comparing with recent imitation learning methods such as Diffusion Policy and Flow Matching~\cite{Rouxel:FlowMatchingMultiContact:Humanoids2024}, as well as incorporating graph-based networks beyond the GCN tested here, to better utilize distributed tactile sensing.



\bibliographystyle{IEEEtran}
\bibliography{main}

\section*{APPENDIX}

\subsection{Details of the Simulation Experiments}

\subsubsection{Simulation Setup}

To enable more extensive evaluation, we constructed a simulation environment using the MuJoCo physics engine. In order to minimize discrepancies between the real-world and simulation results, the simulated robot was configured to closely match the real humanoid robot. The same model-based retargeting and locomotion controller was used in both settings. Distributed tactile sensors were mounted on the robot in the simulator, and tactile signals were simulated via a MuJoCo plugin. The arrangement of sensor cells followed the same hexagonal layout as the e-skin sensor, and the number of patches and cells, as well as the placement on each body part, matched the real robot. In this simulation study, proximity sensing was omitted to avoid excessive computational overhead.

\subsubsection{Manipulation Task}

The task requires the robot to rotate an upright box placed on a table by 90~degrees and lay it down sideways using both arms. As shown in \figref{fig:sim-task}, the box must remain in contact with the table throughout the task, resulting in complex contact transitions. The box is positioned in front of the robot, and we varied its lateral position $p$~[m] across trials to evaluate success rates under different conditions.

\subsubsection{Comparison with Baseline Policies}

\begin{figure}[tpb]
  \centering
  \begin{minipage}[c]{0.48\columnwidth}
    \centering
    \includegraphics[width=0.85\columnwidth]{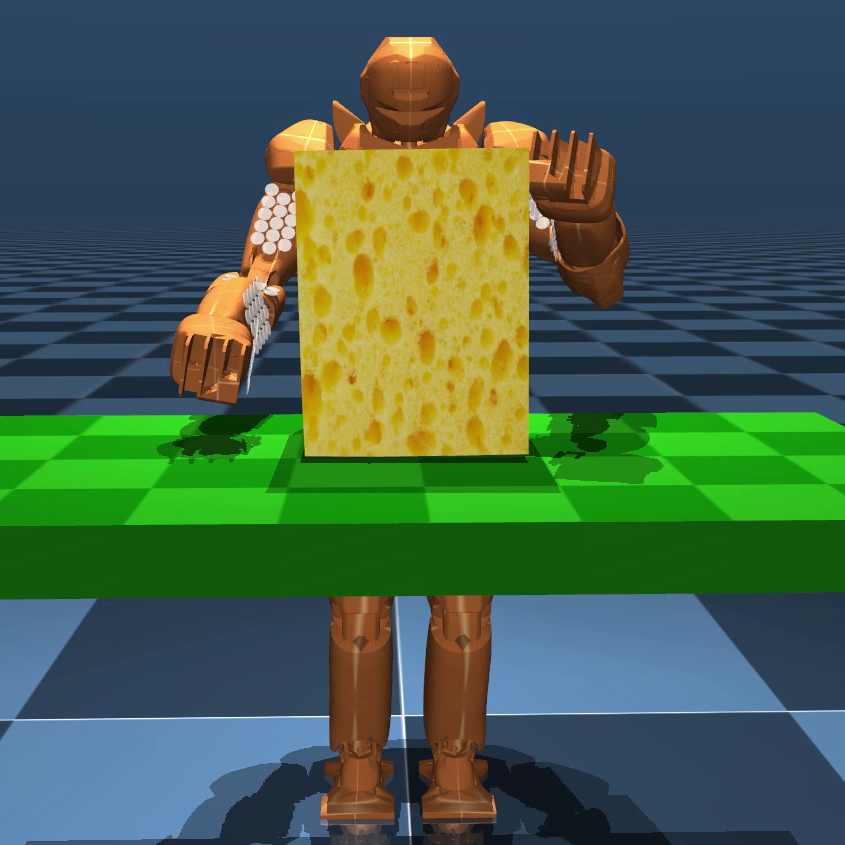}
  \end{minipage}
  \begin{minipage}[c]{0.48\columnwidth}
    \centering
    \includegraphics[width=0.85\columnwidth]{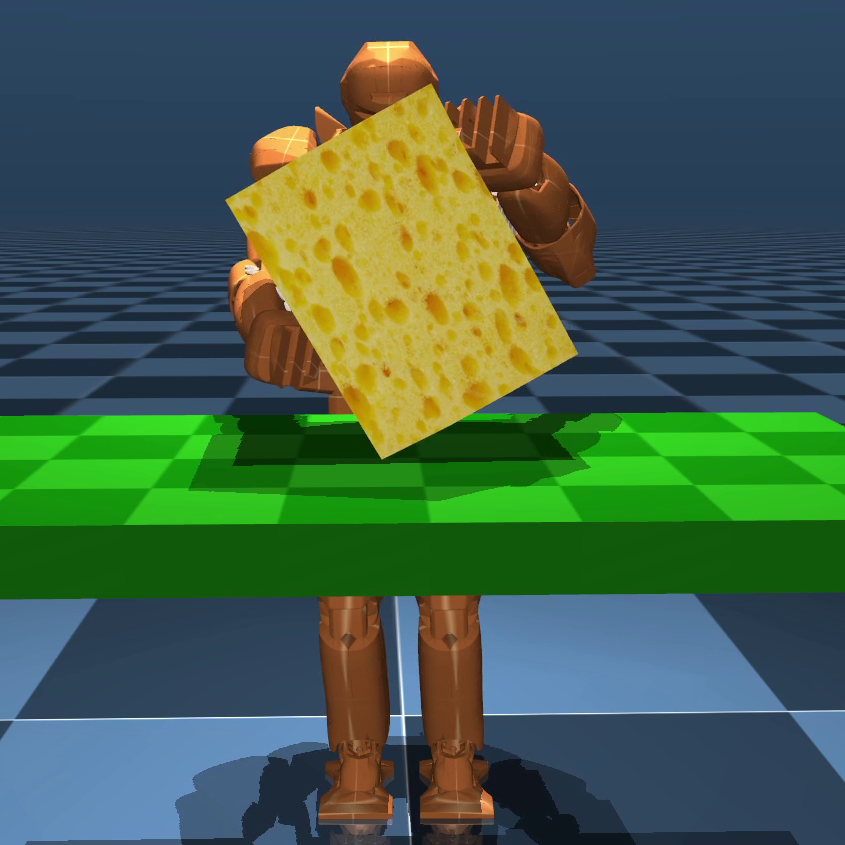}
  \end{minipage}\\
  \vspace{2mm}
  \begin{minipage}{0.48\columnwidth}
    \begin{center} \footnotesize t = 5 s \end{center}
  \end{minipage}
  \begin{minipage}{0.48\columnwidth}
    \begin{center} \footnotesize t = 10 s \end{center}
  \end{minipage}\\
  \vspace{4mm}
  \begin{minipage}[c]{0.48\columnwidth}
    \centering
    \includegraphics[width=0.85\columnwidth]{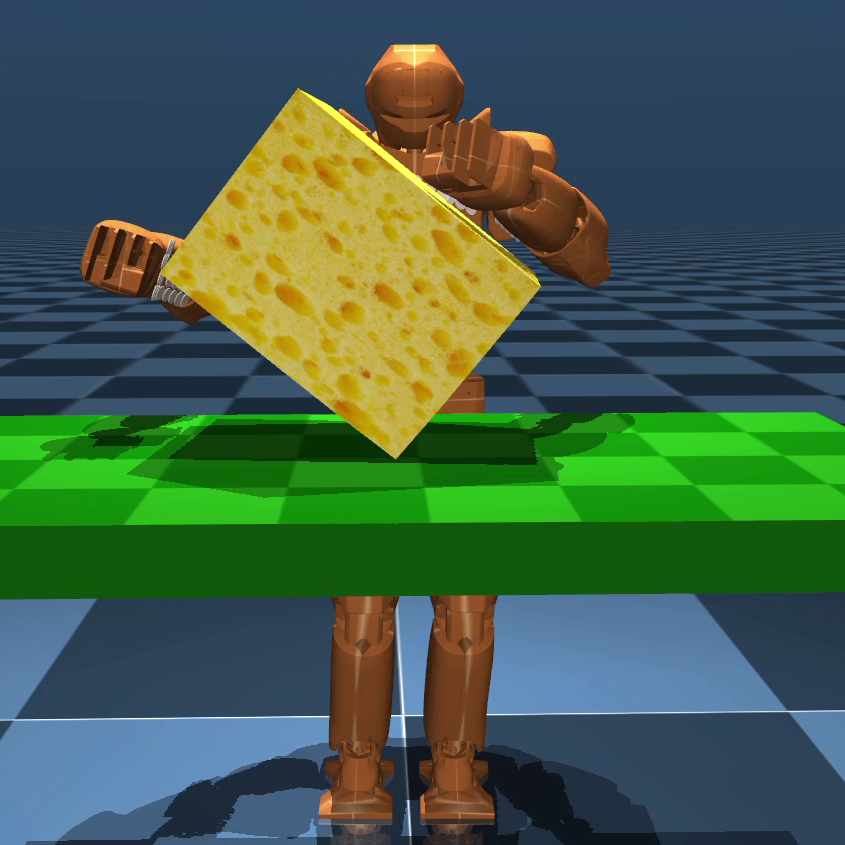}
  \end{minipage}
  \begin{minipage}[c]{0.48\columnwidth}
    \centering
    \includegraphics[width=0.85\columnwidth]{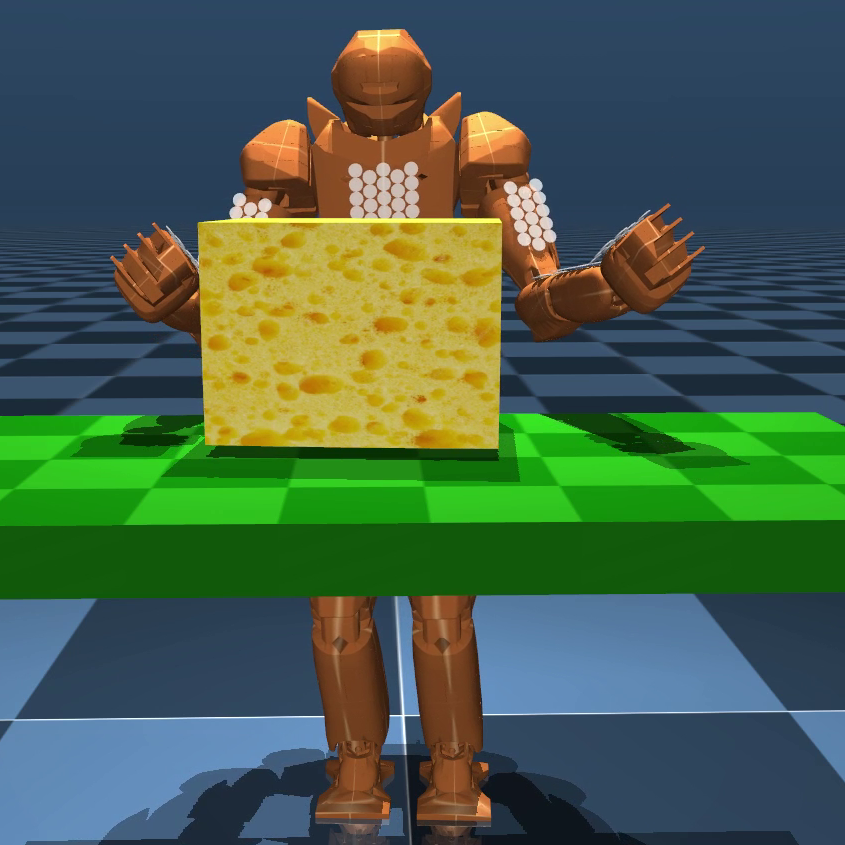}
  \end{minipage}\\
  \vspace{2mm}
  \begin{minipage}{0.48\columnwidth}
    \begin{center} \footnotesize t = 15 s \end{center}
  \end{minipage}
  \begin{minipage}{0.48\columnwidth}
    \begin{center} \footnotesize t = 20 s \end{center}
  \end{minipage}
  \caption{Box reorientation task}
  \label{fig:sim-task}
  \vspace{4mm}
  \centering
  \begin{minipage}[c]{0.44\columnwidth}
    \centering
    \includegraphics[width=1.0\columnwidth]{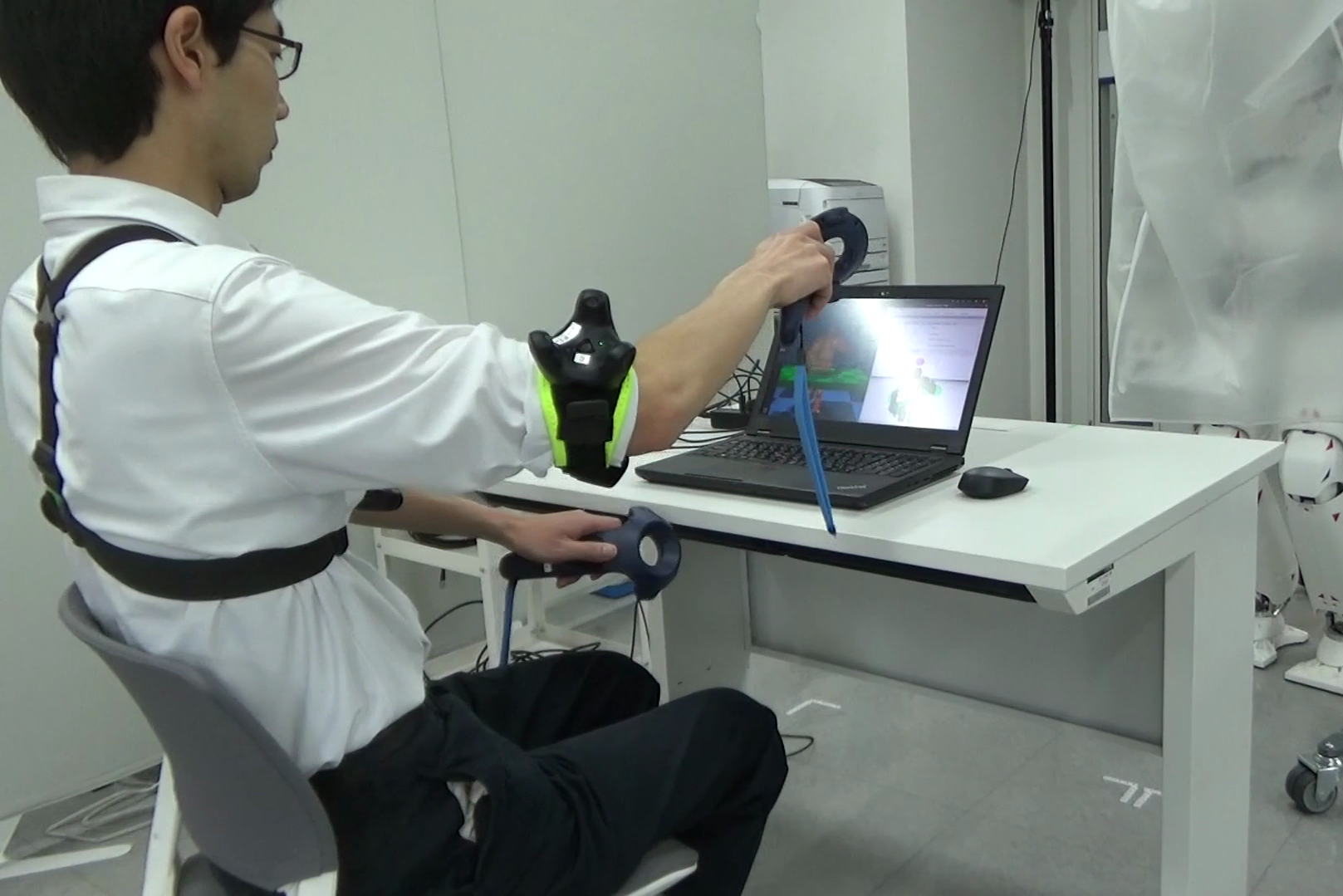}
  \end{minipage}
  \begin{minipage}[c]{0.53\columnwidth}
    \centering
    \includegraphics[width=1.0\columnwidth]{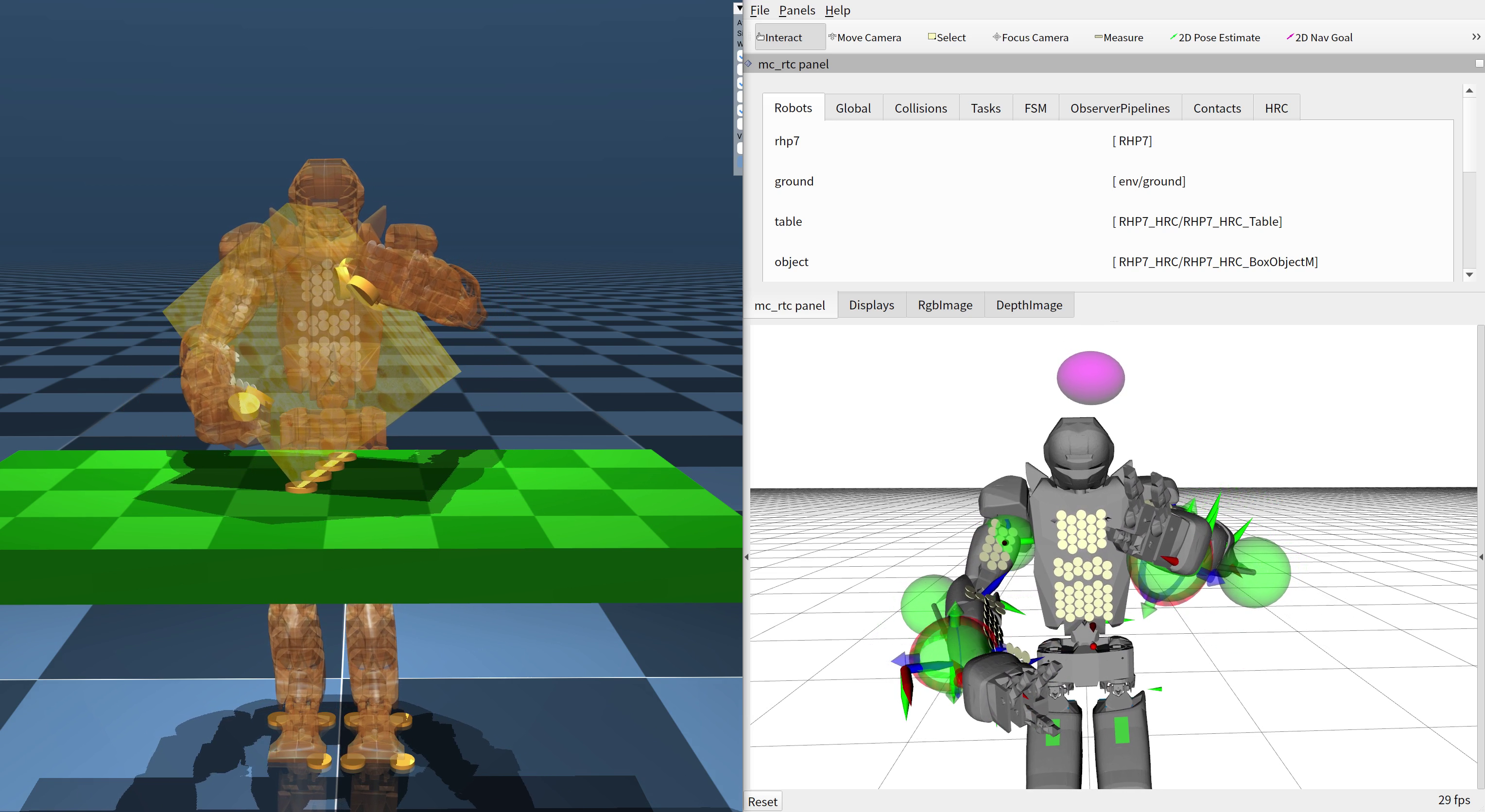}
  \end{minipage}\\
  \vspace{2mm}
  \begin{minipage}{0.48\columnwidth}
    \begin{center} \footnotesize Human operator \end{center}
  \end{minipage}
  \begin{minipage}{0.48\columnwidth}
    \begin{center} \footnotesize Operator display \end{center}
  \end{minipage}
  \caption{Teleoperation via retargeting in the simulation environment.
    \newline
    \footnotesize{
      To support intuitive teleoperation, the manipulated object is rendered semi-transparently, and contact points are visualized with markers in the MuJoCo interface.
  }}
  \label{fig:sim-teleop}
  \vspace{4mm}
  \centering
  \includegraphics[width=0.65\columnwidth]{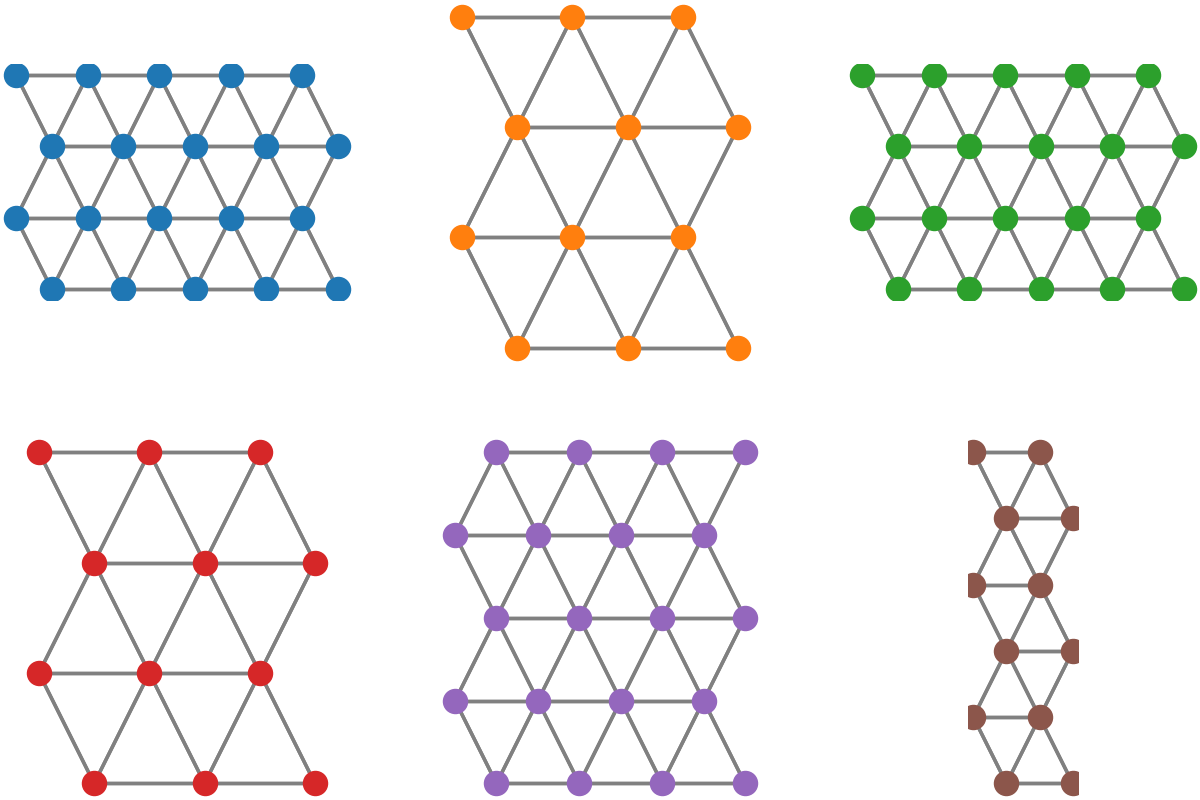}\\
  \caption{Graph structure of tactile sensor cells used in GCN.}
  \label{fig:sim-graph}
\end{figure}

\begin{figure*}[tpb]
  \centering
  \includegraphics[width=0.64\columnwidth]{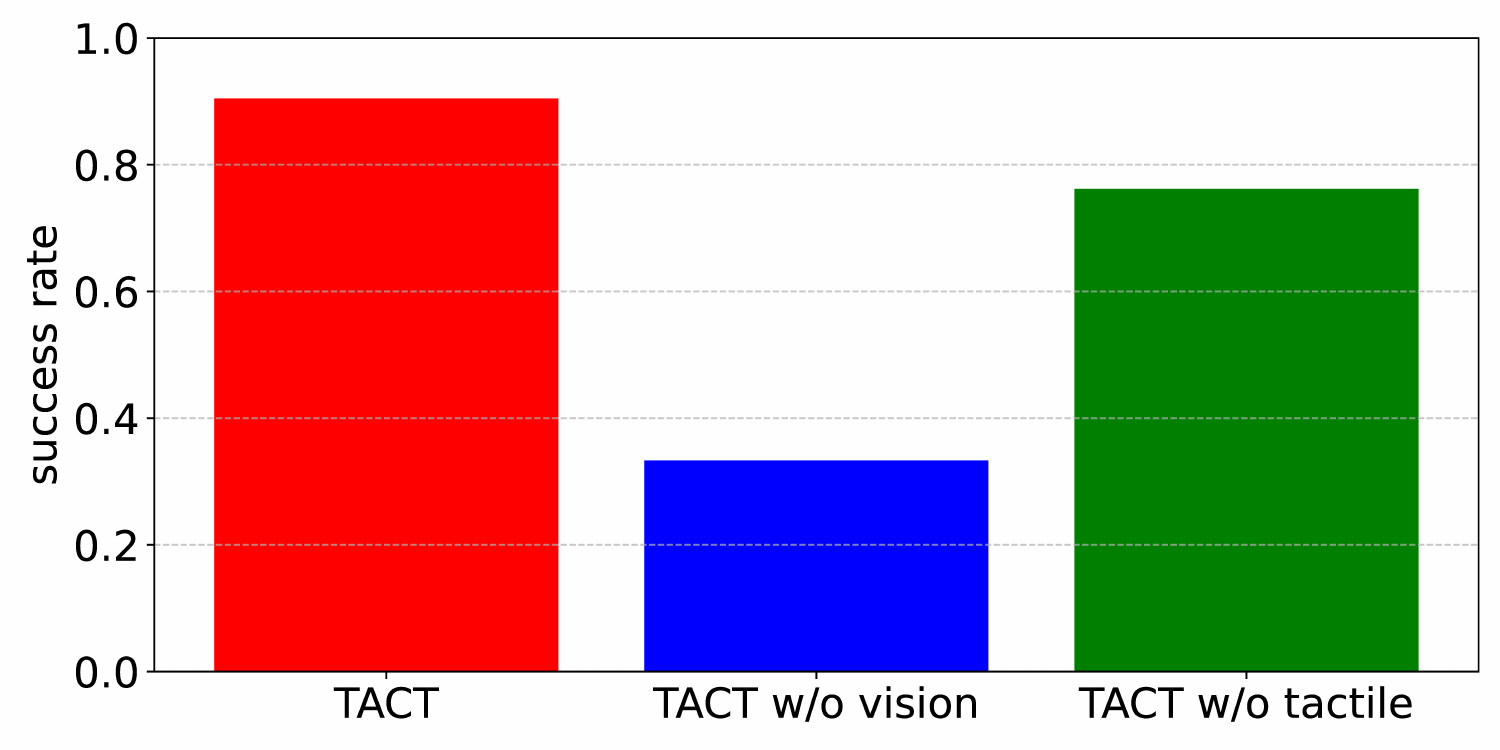}
  \includegraphics[width=0.64\columnwidth]{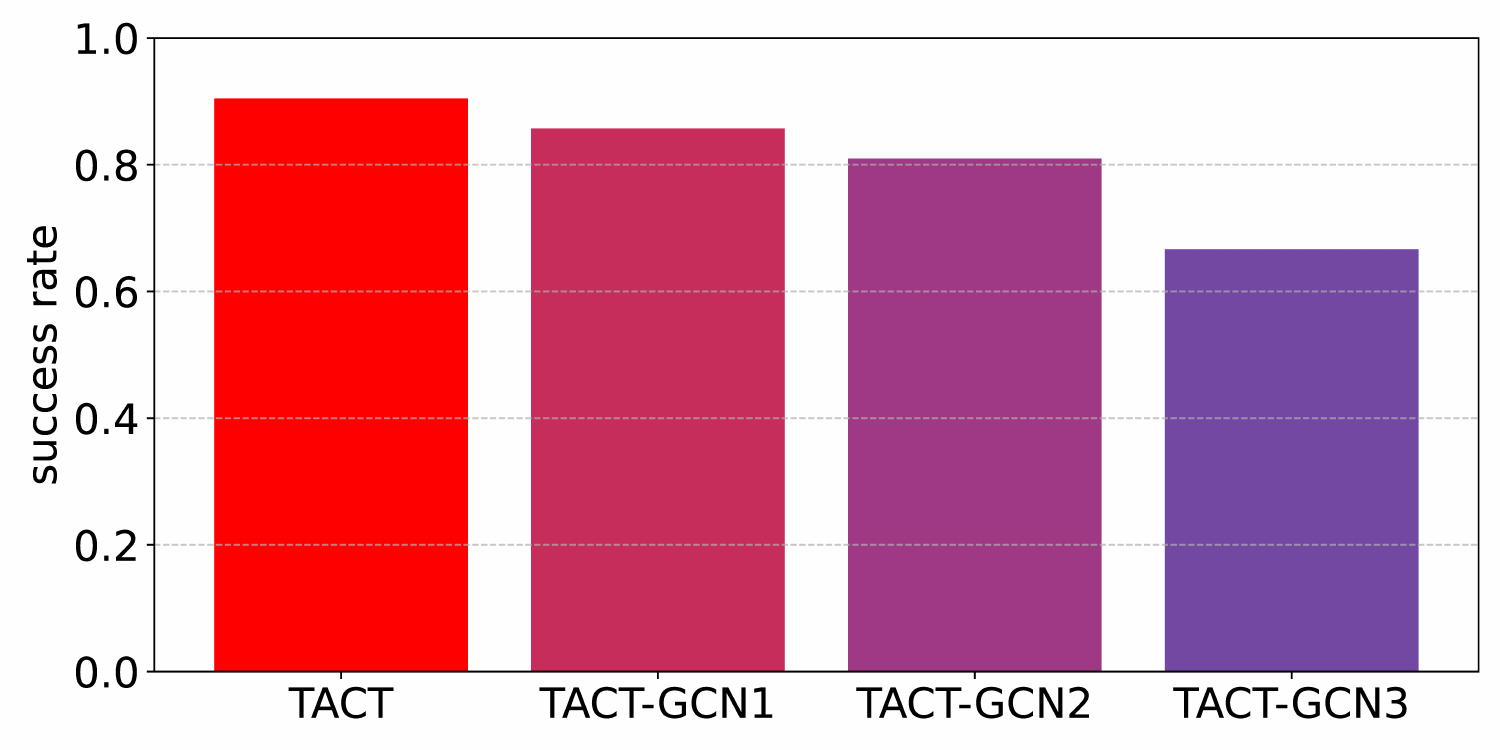}
  \includegraphics[width=0.64\columnwidth]{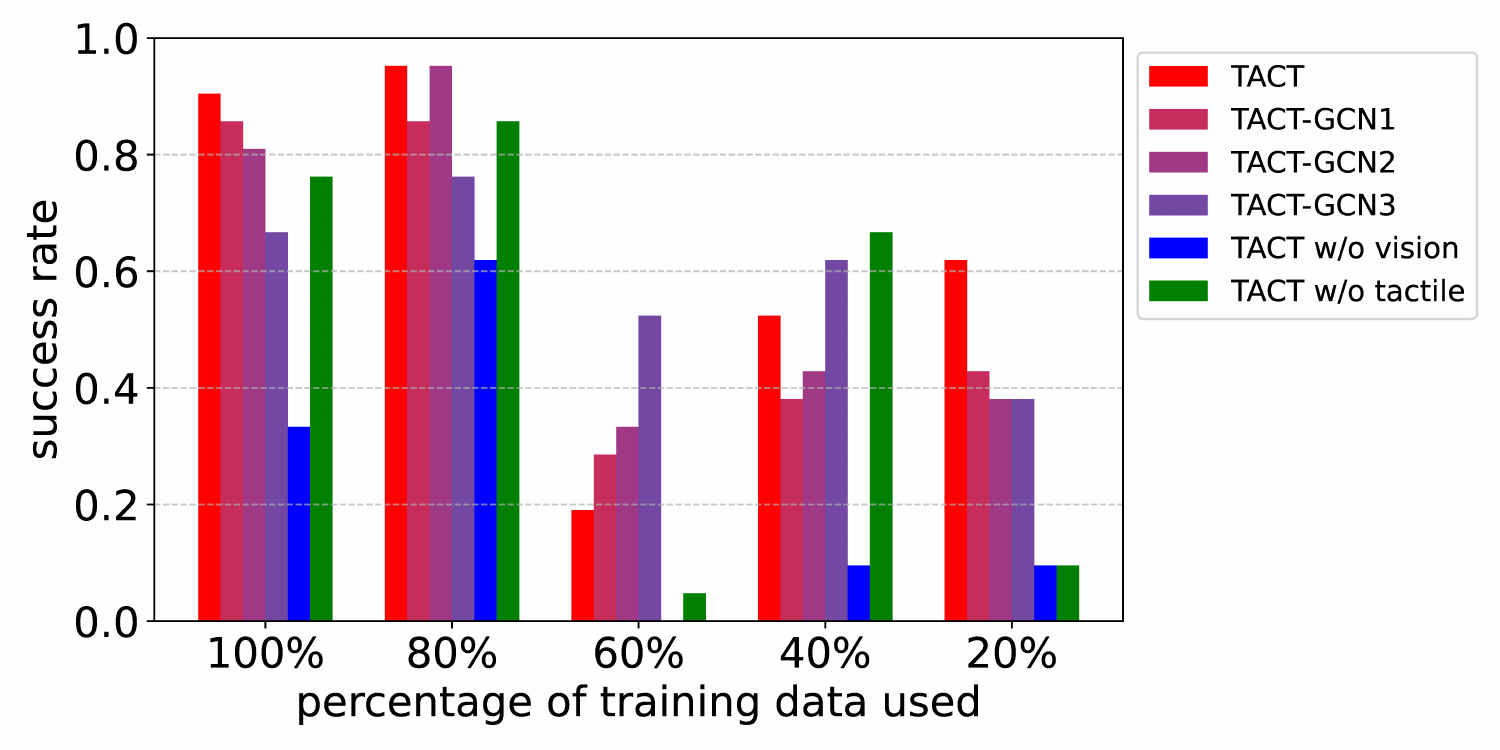}\\
  \begin{minipage}{0.64\columnwidth}
    \begin{center} \footnotesize (A) Baseline comparison \end{center}
  \end{minipage}
  \begin{minipage}{0.64\columnwidth}
    \begin{center} \footnotesize (B) GCN variants \end{center}
  \end{minipage}
  \begin{minipage}{0.64\columnwidth}
    \begin{center} \footnotesize (C) Data efficiency \end{center}
  \end{minipage}\\
  \vspace{2mm}
  \includegraphics[width=0.64\columnwidth]{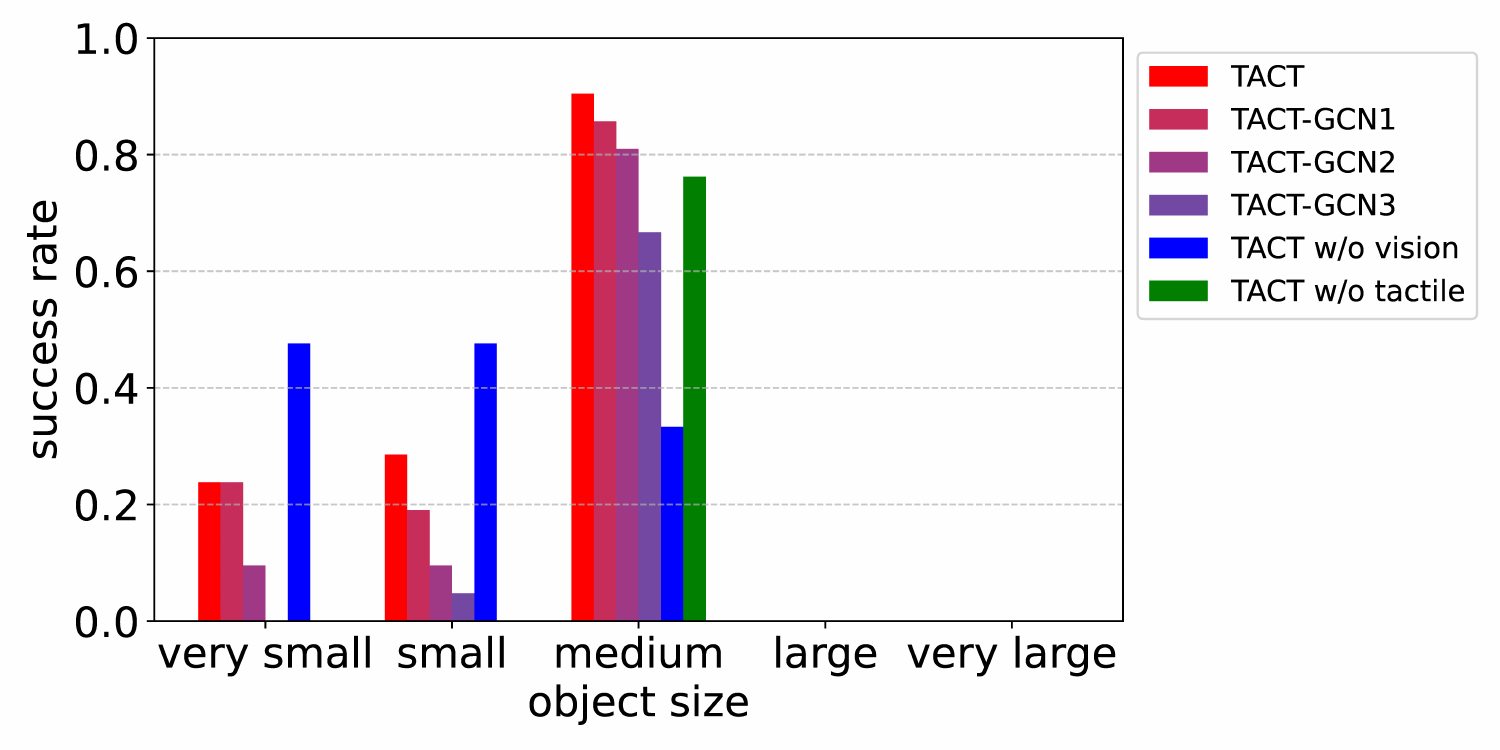}
  \includegraphics[width=0.64\columnwidth]{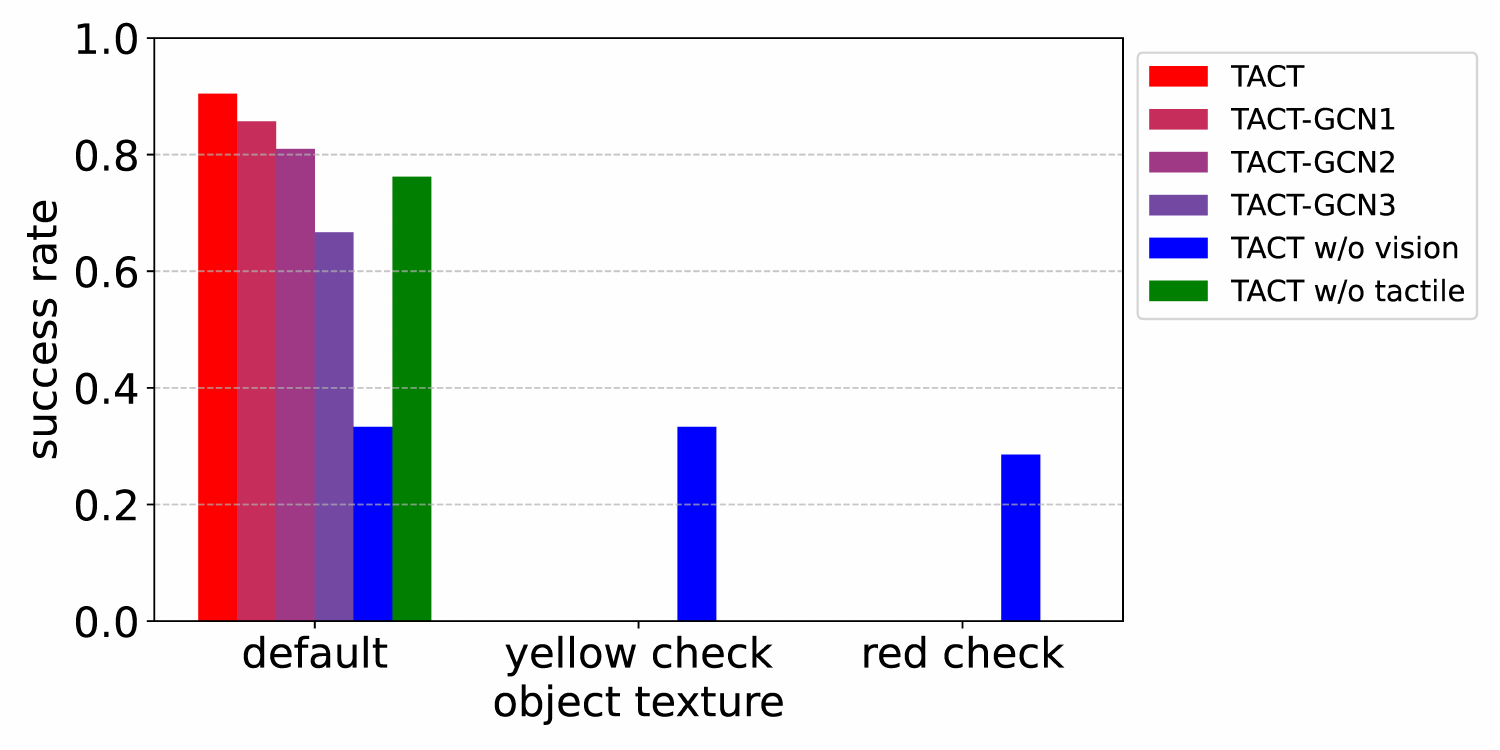}
  \includegraphics[width=0.64\columnwidth]{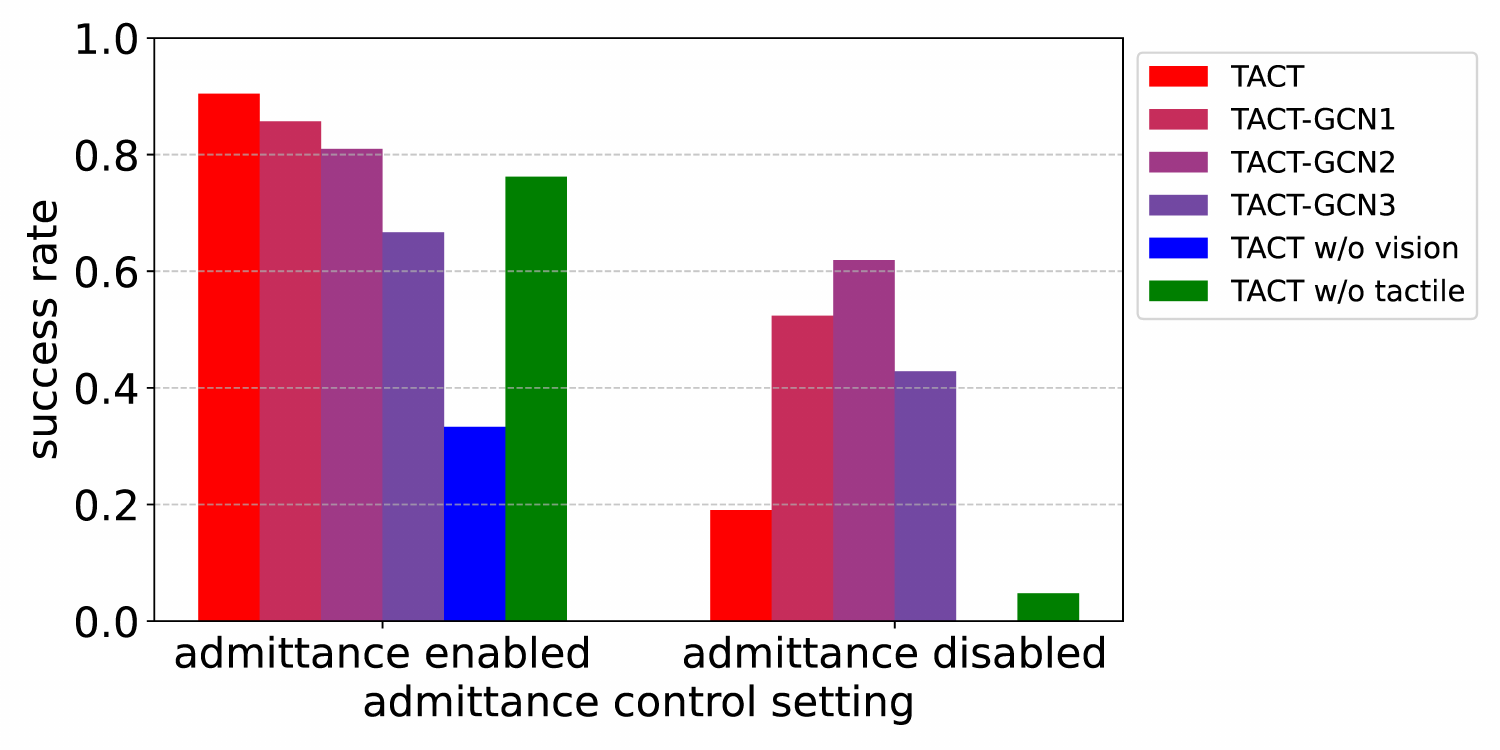}\\
  \begin{minipage}{0.64\columnwidth}
    \begin{center} \footnotesize (D) Object size generalization \end{center}
  \end{minipage}
  \begin{minipage}{0.64\columnwidth}
    \begin{center} \footnotesize (E) Texture generalization \end{center}
  \end{minipage}
  \begin{minipage}{0.64\columnwidth}
    \begin{center} \footnotesize (F) Admittance control \end{center}
  \end{minipage}
  \caption{Success rates for the box reorientation task in simulation experiments.}
  \label{fig:sim-result}
\end{figure*}

As in the real-world experiments, teleoperation data was collected using the same interface as shown in \figref{fig:sim-teleop}. We collected 30 demonstration episodes in total: 10 episodes each for box positions $p = -0.1$, $0.0$, and $0.1$~[m]. Using this dataset, we trained three policies (TACT, TACT w/o vision, and TACT w/o tactile) and evaluated them by deploying each policy at 21 positions ranging from $p = -0.2$ to $0.2$ in 0.02~m increments. As shown in \figref{fig:sim-result}~(A), the proposed TACT policy achieved higher success rates than the baseline policies, consistent with the trends observed in the real-world experiments.
In addition, for each policy, we observed cases where the robot was able to recover from partial failures. Specifically, even when the box initially tilted but subsequently returned to its original orientation, the robot resumed the tilting motion and successfully completed the task. This behavior suggests that the policy acquired a certain level of error recovery capability through imitation learning.

\subsubsection{Incorporating GCNs}

To improve feature extraction from the distributed tactile sensors, we explored the use of graph-based structures. The spatial layout of sensor cells naturally defines a graph, and an adjacency matrix can be constructed based on the hexagonal arrangement. A graph convolutional network (GCN) allows adjacency-aware feature aggregation through linear mappings that consider this structure~\cite{Funabashi:GCNGrasp:RAL2022}. In our implementation, we inserted a graph convolutional layer before the linear layer that generates the tactile token in the TACT architecture.

We evaluated three GCN variants, referred to as TACT-GCN1, TACT-GCN2, and TACT-GCN3, with 1, 2, and 3 graph convolutional layers, respectively. The number of GCN hidden units was set as follows: TACT-GCN1: (16), TACT-GCN2: (16, 32), TACT-GCN3: (16, 32, 64). The adjacency matrix was constructed based on the geometric connectivity of sensor cells, as illustrated in \figref{fig:sim-graph}.

The performance of these variants is shown in \figref{fig:sim-result}~(B). The addition of graph convolutional layers slightly reduced the success rate compared to the original TACT policy, with the degradation becoming more pronounced as the number of layers increased. In subsequent evaluations under different conditions (as presented below), TACT-GCN occasionally outperformed TACT, but the differences remained small. These results suggest that the simple linear projection used in TACT may already be sufficient for feature extraction in our setting, and that incorporating spatial structure via graph-based methods is a promising but non-trivial direction for future work.

\subsubsection{Effect of Reducing Demonstration Data}

We evaluated the data efficiency of the policies by varying the amount of training data. \figref{fig:sim-result}~(C) shows success rates when changing the proportion of the 30 demonstrations used for training. As expected, performance decreases as the training data is reduced. However, several interesting trends emerge.
First, the performance advantage of TACT over the baselines is largely maintained, except at the 40\% level where TACT w/o tactile slightly outperforms TACT. TACT also shows more graceful degradation compared to the baselines.
Second, as the amount of training data decreases, the difference between TACT and TACT-GCN diminishes. At 60\% and 40\% levels, TACT-GCN even outperforms TACT.
Finally, TACT-GCN3 with the deepest graph convolutional layers retains performance better than shallower variants.

\subsubsection{Effect of Object Size Variation}

We tested how well each policy generalizes to different object sizes. \figref{fig:sim-result}~(D) shows the success rates when deploying policies trained only on the original box size to five sizes: very large ($+8$\%), large ($+4$\%), medium (original), small ($-4$\%), and very small ($-8$\%).
Overall, all policies showed significant performance degradation when the object size differed from the training data, highlighting the importance of including size variation during training. Among them, TACT w/o vision (tactile-only) was able to maintain performance even for smaller objects, while TACT w/o tactile (vision-only) failed completely on all non-original sizes. The proposed TACT policy occasionally succeeded with smaller boxes, suggesting that the combination of tactile and visual input offers some degree of generalization beyond the training distribution.

\subsubsection{Effect of Object Texture Variation}

We evaluated how policies generalize to changes in object texture. As shown in \figref{fig:sim-result}~(E), we replaced the original sponge-like texture with yellow and red checkered patterns.
Because all training used a fixed texture, the vision-based policies failed to complete the task with unseen textures. In contrast, TACT w/o vision (tactile-only) was unaffected and maintained its success rate, as expected.

\subsubsection{Effect of Admittance Control}

In the simulation experiments, model-based admittance control, as described in Section~\ref{sec:retargeting}, was enabled by default to improve compliance during contact. \figref{fig:sim-result}~(F) compares performance with and without admittance control. For policies using only vision or tactile input, performance drops sharply when admittance is disabled, likely due to unstable contact transitions. In contrast, TACT and its GCN variants show more resilience, with milder performance degradation. These results indicate that admittance control utilizing tactile feedback plays a complementary role to the policy and should be incorporated to enhance the overall system robustness.

\subsubsection{Robot Falls}

To ensure safety in humanoid loco-manipulation, the robot must avoid falling due to unexpected external forces from the object.
Out of 252 trials, the number of falls per policy was:
TACT: 3, TACT-GCN1: 2, TACT-GCN2: 5, TACT-GCN3: 5, TACT w/o vision: 0, TACT w/o tactile: 10.
TACT w/o tactile exhibited the highest number of falls, while TACT w/o vision showed the lowest. These results suggest that prioritizing tactile input may help reduce the risk of falling during whole-body manipulation.

\end{document}